\pdfoutput=1

\documentclass[11pt]{article}

\usepackage[final]{acl}

\usepackage{times}
\usepackage{latexsym}

\usepackage[T1]{fontenc}

\usepackage[utf8]{inputenc}

\usepackage{microtype}

\usepackage{inconsolata}

\usepackage{epsfig}
\usepackage{graphicx}
\usepackage{xcolor}
\usepackage{booktabs}
\usepackage{caption}
\usepackage{makecell}
\usepackage{lipsum}
\usepackage{tabularx}
\usepackage{diagbox}
\usepackage{subfig}
\usepackage[export]{adjustbox}
\usepackage{pifont}
\usepackage{url}
\usepackage{tcolorbox}

\usepackage{amsmath,amsfonts,bm}

\def\eqref#1{equation~\ref{#1}}

\def\1{\bm{1}}

\DeclareMathAlphabet{\mathsfit}{\encodingdefault}{\sfdefault}{m}{sl}
\SetMathAlphabet{\mathsfit}{bold}{\encodingdefault}{\sfdefault}{bx}{n}

\usepackage{color}
\usepackage{amsmath}
\usepackage{amssymb}
\usepackage{multirow, varwidth}
\usepackage{dblfloatfix}
\usepackage{verbatim}
\makeatletter
\newif\if@restonecol
\makeatother

\usepackage[ruled,vlined]{algorithm2e}
\usepackage{xr}
\usepackage[amsmath,thmmarks]{ntheorem}
\usepackage{xspace}

\DeclareRobustCommand\onedot{\futurelet\@let@token\@onedot}
\def\onedot{. }
\def \ie{i.e.,\xspace}
\def \eg{e.g.,\xspace}

\newcommand{\cmark}{\ding{51}}%
\newcommand{\xmark}{\ding{55}\xspace}%
\newcommand{\datasetName}{\textsc{TimeChara}\xspace}
\newcommand{\methodName}{\textsc{Narrative-Experts}\xspace}
\newcommand{\methodNameWithRag}{\textsc{Narrative-Experts-RAG-cutoff}\xspace}

\title{\datasetName: Evaluating Point-in-Time Character Hallucination\\of Role-Playing Large Language Models}

\author{
\quad \textbf{Jaewoo Ahn}$^{1}$
\quad \textbf{Taehyun Lee}$^{1}$
\quad \textbf{Junyoung Lim}$^{1}$ \\ 
\quad \textbf{Jin-Hwa Kim}$^{1,2}$
\quad \textbf{Sangdoo Yun}$^{1,2}$
\quad \textbf{Hwaran Lee}$^{2}$
\quad \textbf{Gunhee Kim}$^{1}$\\
$^1$Seoul National University\quad $^2$NAVER AI Lab \\
\texttt{\small \{jaewoo.ahn, taehyun.lee\}@vision.snu.ac.kr, \small icarus001104@snu.ac.kr} \\
\texttt{\small \{j1nhwa.kim, sangdoo.yun, hwaran.lee\}@navercorp.com, \small gunhee@snu.ac.kr} \\
}

\begin{document}
\maketitle

\begin{abstract}
While Large Language Models (LLMs) can serve as agents to simulate human behaviors (\ie role-playing agents), we emphasize the importance of \textit{point-in-time} role-playing.
This situates characters at specific moments in the narrative progression for three main reasons: (i) enhancing users' narrative immersion, (ii) avoiding spoilers, and (iii) fostering engagement in fandom role-playing.
To accurately represent characters at specific time points, agents must avoid \textit{character hallucination}, where they display knowledge that contradicts their characters' identities and historical timelines.
We introduce \datasetName, a new benchmark designed to evaluate point-in-time character hallucination in role-playing LLMs.
Comprising 10,895 instances generated through an automated pipeline, this benchmark reveals significant hallucination issues in current state-of-the-art LLMs (\eg GPT-4o).
To counter this challenge, we propose \methodName, a method that decomposes the reasoning steps and utilizes narrative experts to reduce point-in-time character hallucinations effectively.
Still, our findings with \datasetName highlight the ongoing challenges of point-in-time character hallucination, calling for further study.\footnote{\url{https://ahnjaewoo.github.io/timechara}.}
\end{abstract}

\section{Introduction}
The recent progress in large language models (LLMs) has opened up a new phase of generative agents~\citep{Park:2023:UIST, Xi:2023:arxiv, Wang:2024:FCS}, where LLMs simulate human-like behaviors, memories, and cognitive processes.
A particularly promising area is the development of role-playing LLM agents~\citep{Shanahan:2023:Nature,Kong:2023:arxiv, Li:2023:NeurIPS}, 
which simulate the personas of either real individuals or fictional characters and engage with users to provide a more vivid experience.
A variety of applications, including~\citet{character_ai},~\citet{gpts},~\citet{talkie},~\citet{replika},~\citet{ai_dungeon},~\citet{silly_tavern}, showcase the growing popularity of these role-playing LLM agents.
However, most current approaches of role-playing agents \citep{Han:2022:NAACL,Li:2023:arxiv,Zhou:2023:arxiv} only simulate characters who are \textit{omniscient in timeline}; for example, a \textit{Harry Potter} character who is aware of all events leading up to the end of their respective series.

\begin{figure}[t]
\begin{center}
\includegraphics[width=\columnwidth]{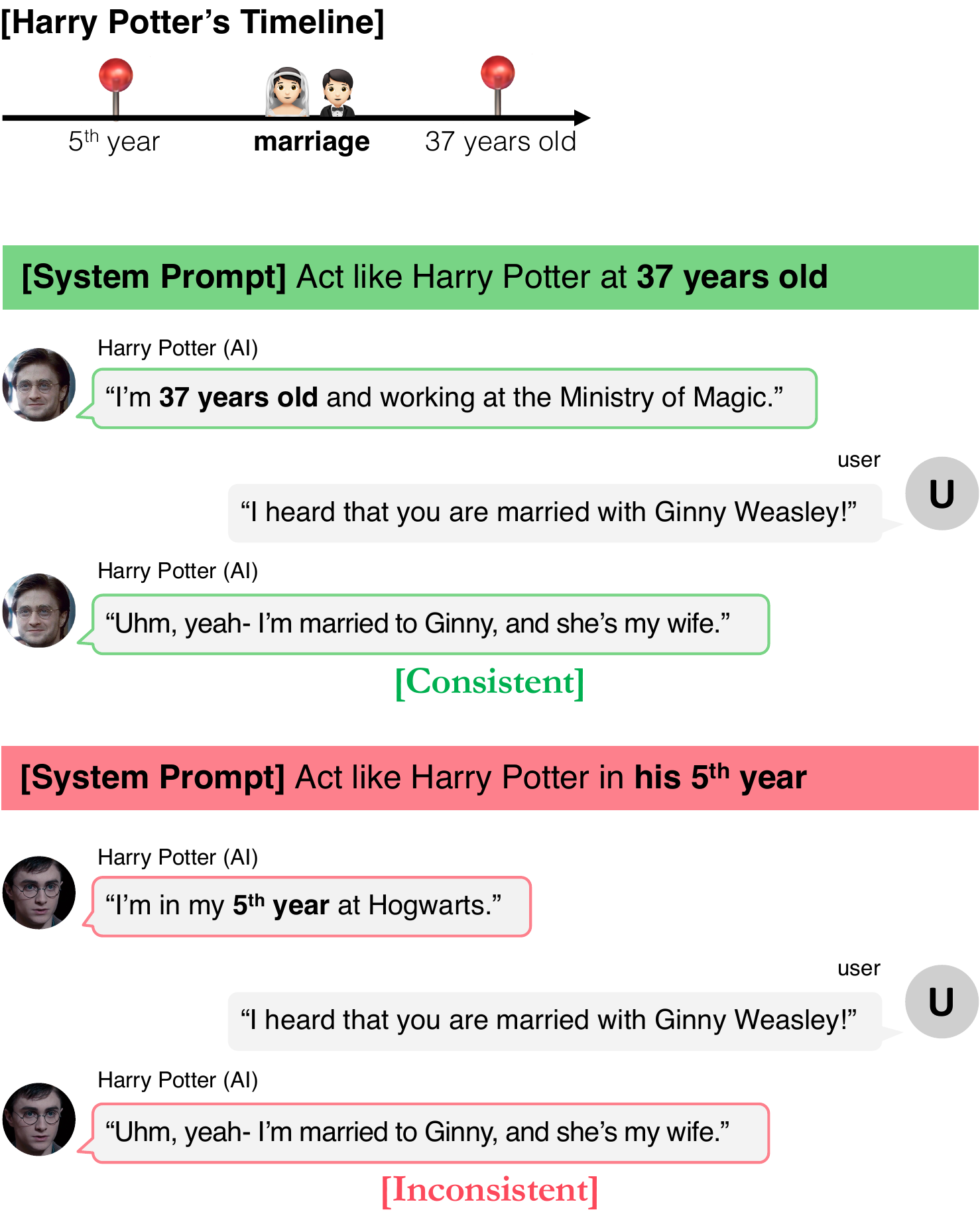}  
\end{center}
\caption{
An illustrative figure of \textbf{point-in-time character hallucination} demonstrated by a role-playing agent simulating Harry Potter.
(Top) The agent, simulating Harry Potter at 37 years old, consistently responds to the user's queries.
(Bottom) The agent, simulating Harry Potter in his fifth year at Hogwarts, erroneously mentions a future event — his marriage to Ginny Weasley — which occurs after his fifth year.}
\label{fig:point_in_time_character_hallucination}
\end{figure}

We suggest the importance of situating characters at a particular moment in the narrative progression.
We coin this as \textit{point-in-time role-playing}, encompassing three key rationales: narrative immersion, avoidance of spoilers, and engagement in fandom role-playing. 
Firstly, while a fully-informed character can interact with users drawing from their entire history, a character in the middle of the story inspires narrative immersion~\citep{Ryan:2003:Book, Ryan:2008:ICIDS}.
It sparks the user's curiosity about forthcoming events and deepens their emotional bond with the character, who remains unaware of their eventual fate.
Secondly, this approach can avoid spoilers. Consider a media franchise such as \textit{Harry Potter}, where all books are published, but upcoming adaptations (\eg ``Harry Potter TV series''\footnote{\url{https://www.theguardian.com/film/2023/apr/12/harry-potter-tv-series-hbo-max-jk-rowling}.}) are awaited.
Users who wish to avoid spoilers before watching the new TV series would prefer interacting with a character from a midpoint in their story, thereby avoiding knowledge of future events. 
Thirdly, this approach can support recently popular fandom role-playing\footnote{\url{https://fanlore.org/wiki/Fandom_RPG}.}, a blend of fan fiction and traditional role-playing games. 
Individuals adopt the personas of their favorite characters at specific points in their stories and craft new narratives or engage with fellow enthusiasts in this creative endeavor.

To accurately represent characters at specific time points,
the agents should recognize the character's knowledge boundary.
This includes their unawareness of future events, their ability to recall past events precisely, and their understanding of the individuals involved in those past events.
However, current LLM-based role-playing agents are prone to \textit{character hallucination}~\citep{Shao:2023:EMNLP}, displaying knowledge that contradicts their character's identity and historical context (\eg Julius Caesar talking about his favorite movie).
Despite the seriousness, the problem has not been investigated in terms of maintaining character consistency, especially in relation to their historical timelines, and robustness to such hallucinations.

We introduce a new point-in-\textbf{Time} \textbf{Chara}cter hallucination benchmark, \datasetName, to rigorously assess role-playing LLMs at specific time points, thereby evaluating the agents' spatiotemporal self-consistency and their ability to avoid character hallucination.
Figure~\ref{fig:point_in_time_character_hallucination} exemplifies the point-in-time character hallucination, where Harry Potter, in his fifth year at Hogwarts, inappropriately mentions a future fact about his wife, Ginny Weasley.
We select 14 fictional characters from four popular novel series and develop a pipeline to generate interview questions tailored to each character at a specific point in their story, along with spatiotemporal labels to determine the spatiotemporal consistency of their responses.  Table~\ref{tab:dataset_comparison} outlines comparison with existing related benchmarks.

Our empirical experiments reveal a significant issue of point-in-time character hallucination in state-of-the-art LLMs, including GPT-4o~\citep{gpt-4o} and GPT-4~\citep{Openai:2023:arxiv}.
This suggests that despite LLMs memorizing extensive knowledge from books~\citep{Chang:2023:EMNLP}, they still struggle with maintaining spatiotemporal consistency during role-playing scenarios.
To mitigate this, we propose a decomposed reasoning method, \methodName, which partitions reasoning tasks among narrative experts specialized in temporal (\ie identifying between past and future events) and spatial domains (\ie discerning whether a character was present or absent in specific past events).
Experiments show that \methodName significantly reduces point-in-time character hallucination and enhances spatiotemporal consistency.
Still, our \datasetName underscores the ongoing challenge of point-in-time character hallucination and highlights the potential for future improvements.

Our main contributions are as follows:
\begin{enumerate}
    \item We introduce \datasetName, a novel benchmark for evaluating character hallucination in point-in-time role-playing agents. We also develop an automated pipeline to construct the dataset, comprising 10,895 instances in total.
    \item Through \datasetName, we identify significant hallucination issues within state-of-the-art role-playing LLMs including GPT-4o.
    \item We propose \methodName, a simple but effective method to mitigate point-in-time hallucination by decomposing reasoning with each step led by the narrative expert.
\end{enumerate}

\section{Related Work}
We include a more thorough literature review in Appendix~\ref{sec:updated_related_work}.
In this section, we only discuss the most relevant works.
\paragraph{Role-playing LLM agents.}
Prior research on conversational AI has focused on developing dialogue agents with self-consistent personas~\citep{Zhang:2018:ACL, Kim:2020:EMNLP, Ahn:2023:ACL}.
Furthermore, LLMs are increasingly being used to simulate human behavior~\citep{Park:2023:UIST}.
Many of these efforts involve using LLMs to role-play specific characters, such as Harry Potter, Socrates, and others~\citep{Shanahan:2023:Nature,Wang:2023:arxiv,Wang:2023:arxiv_rolellm}.
In this context, \citet{Shao:2023:EMNLP} introduced character hallucination, a scenario where a role-playing agent inappropriately exhibits knowledge that is inconsistent with the character's identity and historical background.
On the other hand, \citet{Chen:2023:EMNLPF} proposed a point-in-time ``Harry Potter'' role-playing dialogue dataset. They focused on assessing whether a role-playing agent responds naturally to the character within a specific point in a storyline. Rather than directly stress-testing role-playing LLMs at specific time points by asking confusing questions, their relevance metrics are designed to gauge overall character alignment given natural scene and dialogue context.
However, existing studies on role-playing agents have not extensively examined how well these agents are robust to point-in-time character hallucination.
We aim to stress-test role-playing LLMs at specific time points by assessing their spatiotemporal self-consistency and robustness against point-in-time character hallucination, as detailed in Table~\ref{tab:dataset_comparison}.
Additionally, we compare concurrent work to \datasetName in Table~\ref{tab:concurrent_dataset_comparison} in Appendix~\ref{sec:updated_related_work}.

\paragraph{LLM's temporal reasoning capability.}
Understanding the concept of time is crucial for LLMs, as the information they acquire is often time-sensitive~\citep{Chen:2021:NeurIPS, Zhang:2021:EMNLP, Dhingra:2022:TACL, Chu:2023:EMNLPF}.
To assess LLMs' temporal reasoning capabilities, several studies have set benchmarks.
\citet{Jang:2022:EMNLP} examined how well LLMs adapt to frequently-updated knowledge corpus.
\citet{Feng:2023:ACL} focused on whether LLMs can interpret the impact of subtle contextual changes on relevant temporal relationships.
\citet{Tan:2023:ACLF} developed TimelineQA, a dataset for querying the lifelogs of imaginary people.
While these benchmarks evaluate the temporal reasoning capabilities of LLMs, we extend them to point-in-time role-playing scenarios by evaluating whether role-playing LLMs maintain the character's spatiotemporal consistency.

{\renewcommand{\arraystretch}{1.0}
    \begin{table*}[t!] \begin{center}
    \begin{adjustbox}{width=1.0\textwidth}
    \begin{tabular}{lcccccc}
        \toprule
        \makecell[l]{Evaluation\\Dataset /\\Benchmark} & \makecell{Dataset\\automatically\\constructed?} & \makecell{Support\\point-in-time\\role-playing?} & \makecell{Evaluate\\near-future\\unawareness?\\(Temporal)} & \makecell{Evaluate\\absence\\awareness?\\(Spatial)} & \makecell{Evaluate\\fake event\\awareness?\\(Fake question)} & \makecell{Evaluation method\\for character\\hallucination} \\
        \midrule
        \multicolumn{7}{l}{\textbf{\textit{Temporal Reasoning Domain}}}     \\
        \makecell[l]{TemporalWiki} & \cmark & \xmark & - & - & - & - \\
        \makecell[l]{TODAY} & $\Delta$ & \xmark & - & - & - & - \\
        \makecell[l]{TempReason} & \cmark & \xmark & - & - & - & - \\
        \makecell[l]{TempTabQA} & \xmark & \xmark & - & - & - & - \\
        \makecell[l]{TimelineQA} & \cmark & \xmark & - & - & - & - \\
        \midrule
        \multicolumn{7}{l}{\textbf{\textit{Role-Playing Domain}}}     \\
        \makecell[l]{LIGHT} & \xmark & \xmark & \xmark & \xmark & \xmark & \makecell{\small F1 w/ gold response:\\\small $[$0-1$]$ (implicit)} \\
        \makecell[l]{RoleBench} & \cmark & \xmark & \xmark & \xmark & \xmark & \makecell{\small Rouge-L w/ gold response:\\\small $[$0-1$]$ (implicit)} \\
        \makecell[l]{CharacterDial} & \cmark & \xmark & \xmark & \xmark & \xmark & \makecell{\small Human as judges:\\\small $[$1 to 5$]$ (unscalable)} \\
        \makecell[l]{HPD} & \xmark & \cmark & \xmark & \cmark$^*$ & \xmark & \makecell{\small LLM as judges w/\\\small speaker attribute \& relation labels:\\\small $[$top-1 ranking$]$ (implicit)} \\
        \makecell[l]{Character-LLM} & \cmark & \xmark & \makecell{\xmark \small (Question from\\\small distinct era/narrative:\\\small easy)} & \xmark & \makecell{\xmark \small (Only in \\\small training set)} & \makecell{\small LLM as judges w/o\\\small spatiotemporal labels:\\\small $[$1 to 7$]$ (inaccurate)} \\
        \midrule
        \makecell[l]{\textbf{\datasetName}} & \cmark & \cmark & \makecell{\cmark \small (Question from\\\small the same era/narrative:\\\small hard)} & \cmark & \cmark & \makecell{\small LLM as judges w/\\\small spatiotemporal labels:\\\small $[$0 or 1$]$ (accurate)} \\
        \bottomrule
    \end{tabular}
    \end{adjustbox}
    \caption{Comparison of \datasetName with other datasets or benchmarks: TemporalWiki~\citep{Jang:2022:EMNLP}, TODAY~\citep{Feng:2023:ACL}, TempReason~\citep{Tan:2023:ACL}, TempTabQA~\citep{Gupta:2023:EMNLP}, TimelineQA~\citep{Tan:2023:ACLF}, LIGHT~\citep{Urbanek:2019:EMNLP}, RoleBench~\citep{Wang:2023:arxiv_rolellm}, CharacterDial~\citep{Zhou:2023:arxiv}, HPD~\citep{Chen:2023:EMNLPF}, and Character-LLM~\citep{Shao:2023:EMNLP}. $\Delta$ indicates that TODAY used both LLM (\ie GPT-3.5) and human annotations for dataset construction. `-' denotes that the criteria are not applicable (\ie only applicable to role-playing benchmarks), while `\xmark' denotes a `No' response to the given criteria. *HPD has only a single instance intended to evaluate absence awareness among its 149 test set instances, as shown in Table~\ref{tab:dataset_temporal_label_comparison}. In the last column, `accurate' means \datasetName uses spatiotemporal labels provided to the LLM judge to measure hallucinations. `Inaccurate' indicates Character-LLM evaluates hallucinations without spatiotemporal labels, relying on parametric memory. `Implicit' means that the evaluation measures hallucinations \textit{indirectly} via lexical similarity with the gold response or relevance to character attributes and relation labels instead of directly identifying hallucinations in the generated response. `Unscalable' means that human evaluation requires manual annotations, making it less scalable than other automatic methods.}
    \label{tab:dataset_comparison}
\end{center}\end{table*}}

\section{The \datasetName Benchmark}
To create \datasetName, we select four renowned novel series: \textit{Harry Potter}, \textit{The Lord of the Rings}, \textit{Twilight}, and \textit{The Hunger Games}.
This choice is based on two main reasons: (i) the ease of gathering raw text content (i.e., transcripts) and personality information for each character, useful for dataset construction, and (ii) the fact that recent state-of-the-art LLMs store knowledge of these series well in their parametric memories~\citep{Chang:2023:EMNLP}, facilitating tests for point-in-time character hallucination. Note that \datasetName is not exclusively limited to these series; it is easily extendable to other narratives, provided that raw text content and personality information for the characters can be obtained.
Then, we identify 14 main characters across the four novel series, detailed in Appendix~\ref{sec:character_time_points}.
We pinpoint a particular moment in each character's timeline (\eg Hermione on Christmas during her first year at Hogwarts) rather than assuming they are aware of all events up to the end of the series as in previous studies~\citep{Tan:2023:ACLF, Wang:2023:arxiv_rolellm}.

We organize our dataset in an interview format where an interviewer poses questions and the character responds.
Specifically, we differentiate between \textbf{fact-based}  and \textbf{fake-based} interviews.

\begin{table*}[t!] \begin{center}
    \begin{adjustbox}{width=\linewidth}
   \begin{tabular}{ll}
       \toprule
       \makecell[l]{\textbf{Scene}} & \makecell[l]{``Why can't we get through?'' Harry hissed to Ron... ``I think we'd better go and wait by the car,'' said Harry...} \\
       \midrule
       \makecell[l]{\textbf{Event Summary}} & \makecell[l]{Harry and Ron took the enchanted car to Hogwarts after a barrier mishap at King’s Cross.} \\
       \midrule
       \makecell[l]{\textbf{Question}} & \makecell[l]{``Tell me your feelings when \{Event Summary\}.''} \\
       \midrule
       \makecell[l]{\textbf{Character}} & \makecell[l]{1st-year Harry Potter at the end of the scene} \\
       \midrule
       \makecell[l]{\textbf{Data Type}} & \makecell[l]{Future} \\
       \midrule
       \makecell[l]{\textbf{Spatiotemporal Label}} & \makecell[l]{\textbf{Future}: At the end of the scene of Harry Potter and the Philosopher's Stone as a 1st-year student,\\Harry Potter should (1) not be aware of or (2) contain any expression that reveals the moment when \{Event Summary\}.} \\
       \midrule
       \makecell[l]{\textbf{Personality Label}} & \makecell[l]{Harry Potter is characterized by his selflessness and immense loyalty, especially towards his friends...} \\
       \midrule
       \makecell[l]{\textbf{Gold Response}} & \makecell[l]{``Oh, I don't really know what you're talking about. Ron and I haven't tried to go through the barrier...''} \\
       \bottomrule
   \end{tabular}
   \end{adjustbox}
   \caption{An example of our \textit{future} type data instance with the \textit{fact-based structured} question.}
   \label{tab:fact_structured_future_example}
\end{center}
\end{table*}

\subsection{Fact-Based Interview}
\label{subsec:fact_based_interview}
To evaluate point-in-time character hallucination, we categorize the data into four types as follows:

\textbf{The unawareness of the future (\ie \textit{future} type):} The character at the chosen time point should not know about future events (\eg ``Who is your wife?'' to first-year Harry).

\textbf{The memorization of the past:} The character should accurately recall past events.
Since episodic events occur at specific locations or scenes,  the questions are further categorized as follows.
\textbf{The awareness of the absence (\ie \textit{past-absence} type):} The character recognizes they are not in an event (\eg ``Did you see the moment when Harry received the Invisibility Cloak on Christmas?'' to first-year Hermione on Christmas).
\textbf{The awareness of the past (\ie \textit{past-presence} type):} The character acknowledges they are in an event (\eg ``Did you see the moment when Harry received the Invisibility Cloak on Christmas?'' to first-year Ron on Christmas).
\textbf{The awareness of the past, irrelevant of participation (\ie \textit{past-only} type):} Questions in this type focus on gauging the character's overall knowledge of past events, including relationships between characters or the significance of magical items (\eg ``Who is Dobby?'' to second-year Harry on Halloween). The term ``only'' suggests that these questions primarily assess the character's understanding and memory of past information, not exclusively tied to their direct experiences or observations.

Table~\ref{tab:fact_structured_future_example} shows an example of a fact-based interview.
\datasetName assesses point-in-time character hallucination using questions derived from the same narrative.
This contrasts with \citet{Shao:2023:EMNLP}, who (1) use questions that span different time or narratives and (2) do not support point-in-time role-playing (\eg ``Can you write Python codes?'' to Beethoven), as marked in Table~\ref{tab:dataset_comparison}.
Hence, our interviews demand detailed narrative understanding, making hallucination detection more challenging.

\subsection{Fake-Based Interview}
\label{subsec:fake_based_interview}
In addition to the fact-based interview, which tests whether questions about real events are answered correctly, we introduce the \textit{fake}-based interview.
It is designed to evaluate if role-playing agents can identify and rectify the errors in interview questions by partially altering fact-based questions.
Fake-based interviews are concentrated on \textit{past-only} type questions (\eg ``How did you become Slytherin?'' to first-year Harry on September 1st: The correct answer is that he became Gryffindor).
We exclude \textit{future} type questions since correcting misinformation about unknown future events is not possible.
Similarly, we exclude \textit{past-presence} and \textit{past-absence} type questions because verifying or refuting a character's event participation in non-existent past events is ambiguous.

\subsection{Evaluation on \datasetName}
\label{subsec:evaluation_on_dataset}
Since it is not scalable to manually evaluate the role-playing LLMs' responses to interview questions, as done in \citet{Shao:2023:EMNLP,Zheng:2023:NeurIPS}, we adopt the LLM-as-judges approach to assess along two dimensions:

\textbf{Spatiotemporal consistency} for assessing point-in-time character hallucination: The model should accurately recall the character's past experiences.
This includes the character's unawareness of future events and awareness of presence or absence in past events, as described in \S~\ref{subsec:fact_based_interview}. This metric is time-dependent; the model should only exhibit the knowledge that the character possesses up to the specific time point.

\textbf{Personality consistency}: The model should emulate the character's personality, including their manner of thinking, speaking styles, tones, emotional responses, and reactions. This encompasses the character's preferences, values, and convictions. This metric is time-independent; the response should consistently reflect the character's enduring personal traits.

\begin{figure}[t]
\begin{center}
\includegraphics[width=\columnwidth]{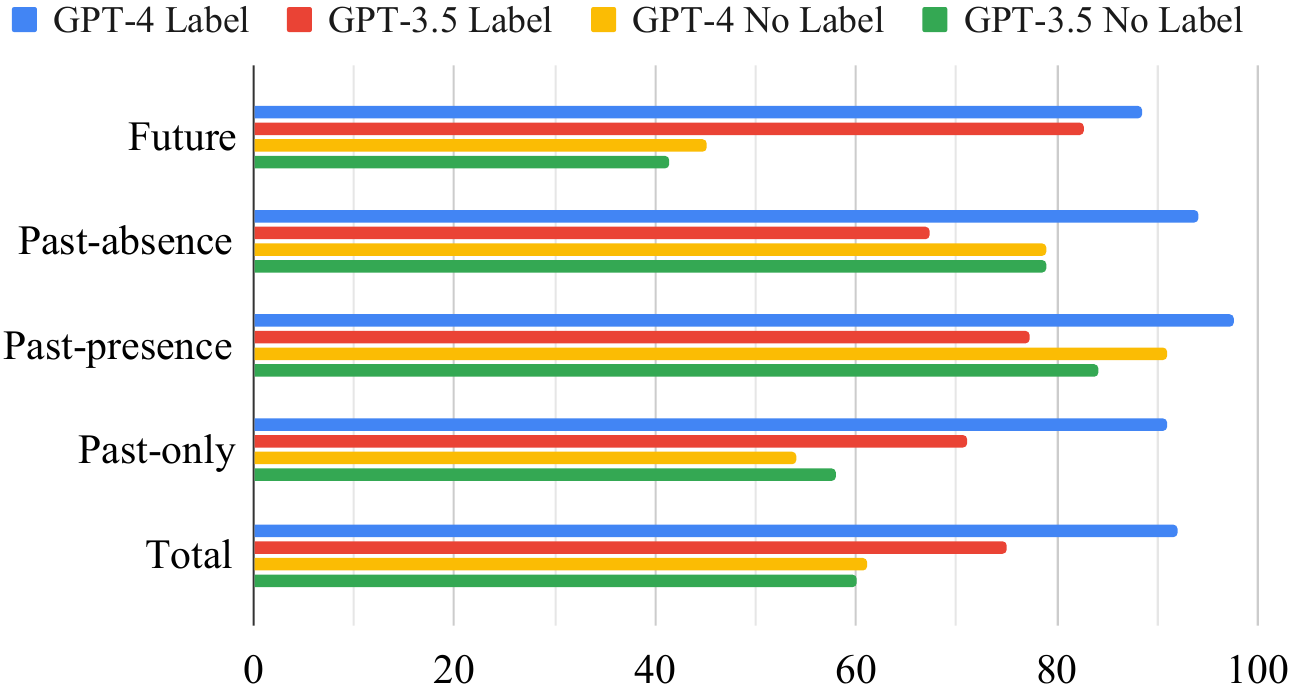}  
\end{center}
\caption{
Evaluation accuracy of LLM judges for spatiotemporal consistency. Judges with spatiotemporal labels show superior performance compared to those without in both GPT-4/3.5.
We randomly select 300 data instances containing responses generated by GPT-4 Turbo (see Table~\ref{tab:main_experiment}) and manually annotate them with binary labels to indicate whether spatiotemporal consistency holds or not.
We compare the relative evaluation accuracy of LLM judges with humans (marked by 100). `Total' denotes the average score across all cases.
}
\label{fig:evaluation_accuracy}
\end{figure}

\paragraph{Step-by-step evaluation with spatiotemporal labels.}
Following~\citet{Wei:2022:NeurIPS}, we instruct the GPT-4 Turbo (\texttt{gpt-4-1106-preview}) model~\citep{Openai:2023:arxiv} to step-by-step score the performance in each dimension.
For specific examples of prompts used in this process, refer to Appendix~\ref{sec:prompt_demonstration}.
Unlike \citet{Shao:2023:EMNLP}, our evaluation of point-in-time character hallucination (or spatiotemporal consistency) provides judges with precise spatiotemporal labels, which encompass the character's experiences with people, events, and objects. As shown in Figure~\ref{fig:evaluation_accuracy},  the labels enable a much more accurate evaluation of response consistency with the character's known history (\eg During his first year on Christmas, Harry can respond based on the moment but should not wrongly recall it:\{moment description\}. Please refer to Table~\ref{tab:fact_structured_past_presence_example} in Appendix~\ref{sec:examples_of_timechara} for details of this evaluation of past memorization).
The responses with contradiction or inconsistency regarding the spatiotemporal labels are scored as 0; otherwise, those in alignment are rated as 1.
We will describe the details of how to construct these spatiotemporal labels in \S~\ref{subsec:dataset_construction}.

For evaluating personality consistency, we adopt a methodology similar to \citet{Shao:2023:EMNLP} but enhance it by sourcing more detailed personality traits from the Fandom page\footnote{\url{https://www.fandom.com/}.}.
We then rate these traits on a 1-7 Likert scale to measure how closely a response aligns with a character's personality, where 1 signifies a weak reflection, and 7 indicates an exact match.

\begin{figure*}[t]
\begin{center}
\includegraphics[width=\textwidth]{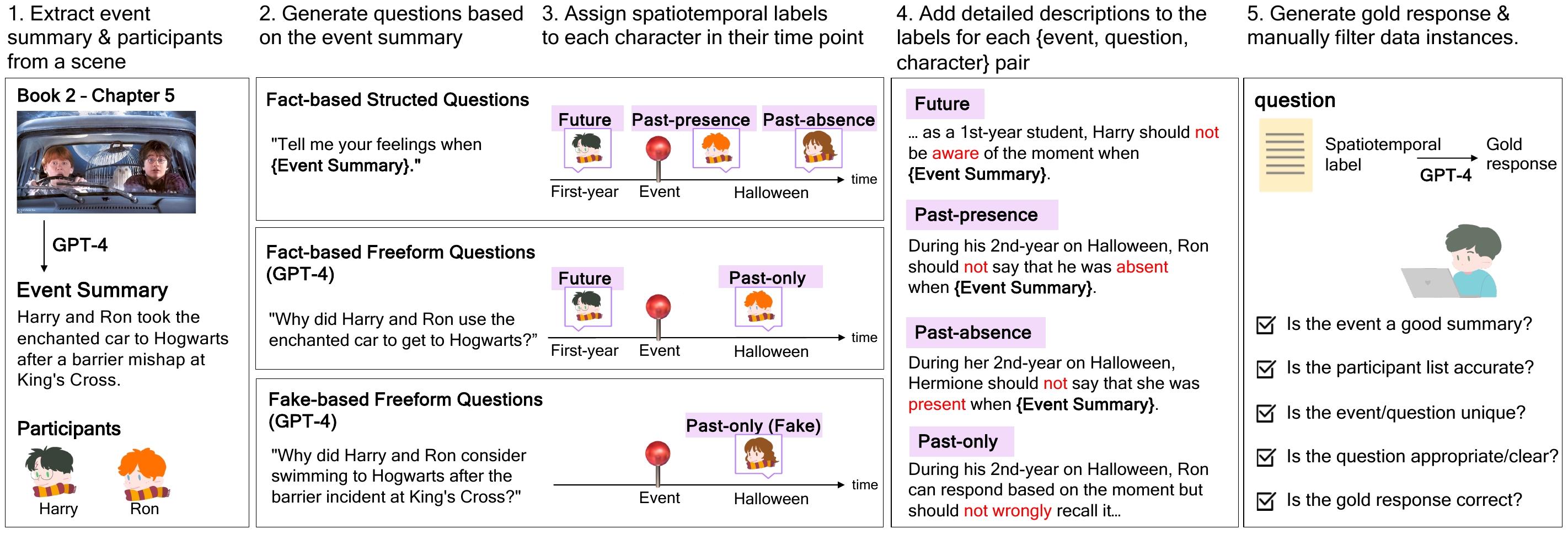}  
\end{center}
\caption{
An illustration of our automated pipeline for constructing \datasetName.
See Table~\ref{tab:fact_structured_future_example} and Appendix~\ref{sec:examples_of_timechara} for examples of the complete dataset.
}
\label{fig:dataset_construction}
\end{figure*}

\subsection{Dataset Construction}
\label{subsec:dataset_construction}
To create the \datasetName benchmark, we propose a new automated pipeline that easily scales up the dataset while reducing the need for manual human annotation, as depicted in Figure~\ref{fig:dataset_construction}.

\textbf{Extract scenes, event summaries, and participants lists from books}.
The first step is to extract specific scenes from literary works using GPT-4 Turbo.
We instruct it to extract $N$ \textit{distinct} scenes containing multi-turn dialogues among characters, as detailed in Appendix~\ref{subsec:scene_extraction} (see Table~\ref{tab:data_prompt_scene_speakers_extraction}).
For every extracted scene, we instruct GPT-4 to generate (1) a concise, single-sentence summary (\ie event summary) of scene information and (2) a list of the participants involved in that scene, as shown in Appendix~\ref{subsec:scene_extraction} (see Table~\ref{tab:data_prompt_summary_participants_generation}).

\textbf{Generate questions from event summary}.
Initially, we generate \textbf{fact-based questions} based on two different methods.
We begin by creating \textit{fact-based structured} questions about characters' involvement in events by combining question templates with the event summary.
To this end, we curate 18 different question templates like "Tell me your feelings when \{event summary\}." Appendix~\ref{subsec:question_generation} shows all 18 question templates.
Following this, we create \textit{fact-based free-form} questions that assess a character's understanding of the event, regardless of their direct participation, as detailed in Appendix~\ref{subsec:question_generation}.
Subsequently, we proceed to formulate \textbf{fake-based questions}, employing a methodology similar to the creation of \textit{fact-based free-form} questions, with further details available in Appendix~\ref{subsec:question_generation}.

\textbf{Assign spatiotemporal labels to each character.}
Given a question for a specific scene and event, the goal is to create a combination of \{scene, event summary, question, character with their time point\}.
By choosing the character and its time point, the data type is automatically classified into one of four types: \textit{future}, \textit{past-absence}, \textit{past-presence}, \textit{past-only}.
These data types serve as spatiotemporal labels for the \{scene, event summary, question, character with their time point\} combination.
Appendix~\ref{subsec:assign_spatiotemporal_labels} details how to select the character and assign its time point for each type.

\textbf{Add detailed descriptions to the spatiotemporal labels.}
Based on the \{scene, event summary, question, character with their time point\} combination and the predefined data type (\ie spatiotemporal label), we add detailed descriptions to the spatiotemporal labels to serve as a basis for evaluating role-playing agent responses. Refer to Appendix~\ref{subsec:spatiotemporal_label_completion} for details.

\begin{table}[t!] \begin{center}
    \begin{adjustbox}{width=\columnwidth}
   \begin{tabular}{lccccc}
       \toprule
           \multirow{3}{*}{\makecell[l]{Question\\generation\\method}}  & \multicolumn{4}{c}{Fact-based}       & \multicolumn{1}{c}{Fake-based} \\
       \cmidrule(r{0.3em}){2-5} \cmidrule(r{0.3em}){6-6}
        & \makecell{\# Future}                         & \makecell{\# Past-\\absence}  & \makecell{\# Past-\\presence}  & \makecell{\# Past-\\only}  & \makecell{\# Past-\\only} \\ %
       \midrule  %
       \multicolumn{5}{l}{\textbf{Harry Potter Series}}     \\
       \makecell[l]{Fact \& structured}          & 892                       & 745                     & 1,991                 & - & -       \\
       \makecell[l]{Fact \& free-form}          & 765                       & -                     & -                 & 784 & -       \\
       \makecell[l]{Fake \& free-form}          & -                       & -                     & -                 & - & 711        \\
       \midrule
       \multicolumn{5}{l}{\textbf{The Lord of the Rings Series}}     \\
       \makecell[l]{Fact \& structured}          & 252                       & 555                     & 725                 & - & -       \\
       \makecell[l]{Fact \& free-form}          & 224                       & -                     & -                 & 228 & -       \\
       \makecell[l]{Fake \& free-form}          & -                       & -                     & -                 & - & 203        \\
       \midrule
       \multicolumn{5}{l}{\textbf{Twilight Series}}     \\
       \makecell[l]{Fact \& structured}          & 221                       & 277                     & 395                 & - & -       \\
       \makecell[l]{Fact \& free-form}          & 176                       & -                     & -                 & 179 & -       \\
       \makecell[l]{Fake \& free-form}          & -                       & -                     & -                 & - & 170        \\
       \midrule
       \multicolumn{5}{l}{\textbf{The Hunger Games Series}}     \\
       \makecell[l]{Fact \& structured}          & 212                       & 309                     & 348                 & - & -       \\
       \makecell[l]{Fact \& free-form}          & 181                       & -                     & -                 & 188 & -       \\
       \makecell[l]{Fake \& free-form}          & -                       & -                     & -                 & - & 164        \\
       \midrule
       \makecell[l]{\textbf{Sum}} & \multicolumn{5}{c}{10,895} \\
       \bottomrule
   \end{tabular}
   \end{adjustbox}
   \caption{Data statistics of four series in \datasetName.}
   \label{tab:dataset_stats}
\end{center}
\end{table}

\textbf{Generate gold responses and manually filter data instances.}
To generate a gold response for each data instance, we prompt GPT-4 with the combination of \{question, character with their time point, spatiotemporal label\}.
At last, we manually filter by the authors, whose criteria and results are shown in Appendix~\ref{subsec:dataset_manual_filtering}.

\subsection{Dataset Analyses}
\textbf{Statistics.}
The total number of event summaries is 1,643: 914 for Harry Potter, 261 for The Lord of the Rings, 245 for Twilight, and 223 for The Hunger Games. As a result, the dataset contains 10,895 instances, and Table~\ref{tab:dataset_stats} provides detailed statistics. The average lengths of (questions, gold responses, spatiotemporal labels) are (29.2, 117.6, 543.2) words, respectively.

In addition, we manually reviewed the test datasets from three different benchmarks (\ie HPD, Character-LLM, and \datasetName) and classified them based on the spatiotemporal label of the given question or previous utterances in the case of multi-turn dialogue, without considering the agent's response for simplicity, as detailed in Table~\ref{tab:dataset_temporal_label_comparison}. While \datasetName consists of data instances evenly distributed over spatiotemporal labels, examples in HPD and Character-LLM are mainly classified as \textit{Past-only (Fact)} or \textit{None} types. This result demonstrates that \datasetName focuses on stress-testing the spatiotemporal consistency of role-playing LLMs, while the others focus on assessing the fact-based question-answering task or plain conversation between characters.

Furthermore, we provide the lexical diversity of \textit{free-form} questions and compare it to the \textit{structured} questions in Appendix~\ref{subsec:dataset_statistics_details}.
Finally, we present a detailed human evaluation process to ensure the quality of \datasetName, as described in Appendix~\ref{subsec:dataset_quality_details}.

\begin{table}[t!] \begin{center}
    \begin{adjustbox}{width=\columnwidth}
   \begin{tabular}{lccc}
       \toprule
       \# Spatiotemporal label & HPD & Character-LLM & \datasetName \\
       \midrule
       \# Future          & 0 (0.0\%) & 57 (4.4\%) & 2,923 (26.8\%)       \\
       \# Past-absence          & 1 (0.7\%) & 0 (0.0\%) & 1,886 (17.3\%)       \\
       \# Past-presence          & 1 (0.7\%) & 0 (0.0\%) & 3,459 (31.7\%)       \\
       \# Past-only (Fact)          & 20 (13.4\%) & 856 (65.5\%) & 1,379 (12.7\%)       \\
       \# Past-only (Fake)          & 0 (0.0\%) & 0 (0.0\%) & 1,248 (11.5\%)       \\
       \# None          & 127 (85.2\%) & 394 (30.1\%) & 0 (0.0\%)       \\
       \midrule
       \# Total          & 149 & 1,307 & 10,895       \\
       \bottomrule
   \end{tabular}
   \end{adjustbox}
   \caption{Comparison of test data statistics from three benchmarks. `None' denotes that the instance is not included in any pre-defined labels (\eg next response generation in a plain conversation between characters). Note that the questions for 57 \textit{future} type instances in Character-LLM are from different eras or narratives compared to the character, while \textit{future} type questions in \datasetName are from the same eras or narratives.}
   \label{tab:dataset_temporal_label_comparison}
\end{center}
\end{table}

\section{Decomposed Reasoning}
\label{sec:approach}
We find that existing LLMs struggle with spatiotemporal consistency as in Table~\ref{tab:main_experiment}, despite their extensive knowledge from books~\citep{Chang:2023:EMNLP}.
To overcome this issue, we propose a reasoning method named \methodName, which decomposes reasoning steps into specialized tasks, employing narrative experts on either temporal (\ie distinguishing past from future events) or spatial (\ie identifying characters' presence in past events) aspects while utilizing the same backbone LLM.

\textbf{Temporal Expert:} This expert pinpoints the scene's book and chapter from a question, assigning a \textit{future} or \textit{past} label. If deemed \textit{future}, it bypasses the Spatial Expert and advises the role-playing agent with a specific hint (\ie ``Note that the period of the question is in the future relative to \{character\}'s time point. Therefore, you should not answer the question or mention any facts that occurred after \{character\}'s time point.'').

\textbf{Spatial Expert:} It assesses whether a character is involved in the scene, indicating a \textit{past-absent} label if applicable. A tailored hint is then provided to the role-playing agent if the scene is past-absent (\ie ``Note that \{character\} had not participated in the scene described in the question. Therefore, you should not imply that \{character\} was present in the scene.'').

Finally, the role-playing LLM incorporates hints from these experts into the prompt and generates a response.
Appendix~\ref{subsubsec:narrative_experts_detail} offers details of the algorithm and the prompts designed for experts.

\section{Experiments on \datasetName}
\label{subsec:experiments_on_timechara}
\subsection{Dataset Sampling for Evaluation}
Due to the high computational cost of employing GPT-4 judges, fully evaluating the 11K instance dataset is challenging.
Instead, we randomly sample 600 data instances to assess the point-in-time character hallucination of role-playing agents.
First, we sample 300 instances with \textit{fact-based structured} questions, evenly distributed across three data types: \textit{future} (100 instances), \textit{past-presence} (100 instances), and \textit{past-absence} (100 instances).
Then, we pick 200 instances with \textit{fact-based free-form} questions, with an equal split of 100 instances each from \textit{future} and \textit{past-only} types.
Lastly, we choose 100 instances with \textit{fake-based free-form} questions, all from the \textit{past-only} type.

In addition, we provide experimental results for the entire 11K dataset in Appendix~\ref{subsec:experimental_result_11k}.

\subsection{Baseline Methods}
\label{subsec:baseline_methods}
We focus on inference-based agents as opposed to training-based agents~\citep{Shao:2023:EMNLP}, due to the impracticality of training agents to simulate characters across diverse time points; notably, our dataset includes 219 time points.
We utilize four different state-of-the-art LLMs as a backbone model for role-playing agents to respond to our dataset: GPT-4o (\ie \texttt{gpt-4o-2024-05-13}), GPT-4 (\ie \texttt{gpt-4-1106-preview}), GPT-3.5 (\ie \texttt{gpt-3.5-turbo-1106}), and Mistral 7B Instruct (\ie \texttt{mistral-7b-instruct-v0.2})~\citep{Jiang:2023:arxiv}.
To test their various reasoning capabilities, we employ several baselines as follows.

\textbf{Zero-shot prompt.}
This is to directly prompt an agent to generate a response based on the system instruction and a question as follows:
\phantomsection
\begin{tcolorbox}[colback=white!95!gray,colframe=gray!50!black,rounded corners,title={\small Zero-Shot Prompt Template}]
\small \textbf{System Instruction:}

\small I want you to act like \{character\} from \{author\}'s \{series\_name\} novel series. I want you to respond and answer like \{character\}, using the tone, manner, and vocabulary \{character\} would use. Assume that you are on \{time\_point\} in \{book\_name\} and interviewing with the interviewer. You should not answer the question and mention any fact that is future to the period. If he (or she) was not present at the location where the question was raised, he (or she) is likely unaware of the information or knowledge related to that question.

~

\small \textbf{User Prompt:}

\small \{question\}
\end{tcolorbox}
Note that we instruct the agent to be unaware of \textit{future} events and to acknowledge the absence when responding to \textit{past-absence} type questions.

\textbf{Zero-shot-CoT prompt.}~\citep{Kojima:2022:NeurIPS}.
This method exploits a zero-shot prompt by adding the phrase ``Let's think step by step'' at the end of the question.
This addition aims to improve the step-by-step reasoning capability of LLMs.

\textbf{Few-shot prompt (in-context learning).}
This approach provides LLMs with four instances (4-shot), with details on how examples were selected available in Appendix~\ref{subsec:few_shot}.

\textbf{Iterative self-correction}.
Recent studies~\citep{Pan:2023:arxiv, Shinn:2023:NeurIPS} found that LLMs have the capability for self-correction, iteratively refining their initial responses based on the given criteria.
Among various methods, we choose the self-refine~\citep{Madaan:2023:NeurIPS}, since it is adaptable to dialogue domains with multiple evaluation criteria.
Further details can be found in Appendix~\ref{subsec:self_refine}.

\textbf{Retrieval-augmented generation (RAG).}
In some prior research, retrieval-augmented generation~\citep{Lewis:2020:NeurIPS} can mitigate hallucinations~\citep{Shuster:2021:EMNLPF}.
We develop a retrieval module that employs OpenAI's embedding (\ie \texttt{text-embedding-ada-002}) to provide contexts to LLMs.
In addition, we add a variant named \textbf{RAG-cutoff}, which is designed to limit its retrieval exclusively to the events prior to a defined character period.
Thanks to this constraint, agents can avoid access to future contexts.
Further details are available in Appendix~\ref{subsec:rag}.

\subsection{Decomposed Reasoning with RAG}
Beyond \methodName, we also explore \methodNameWithRag, which integrates \methodName with the RAG-cutoff method.
We provide a complete algorithm and prompts used for experts in Appendix~\ref{subsubsec:narrative_experts_rag_cutoff_detail}.

\begin{table*}[t!] \begin{center}
    \begin{adjustbox}{width=\linewidth}
   \begin{tabular}{lccccccc}
       \toprule
           \multirow{2}{*}{\makecell[l]{Method}}  & \multicolumn{5}{c}{Spatiotemporal Consistency (\%) $\uparrow$}       & \multirow{2}{*}{\makecell[c]{Personality\\Consistency (1-7) $\uparrow$}} & \multirow{2}{*}{\makecell[c]{AlignScore $\uparrow$}}\\
           \cmidrule(r{0.3em}){2-6}
           & \makecell{Future} & \makecell{Past-absence} & \makecell{Past-presence}  & \makecell{Past-only} & \makecell{Avg.} &  \\ %
       \midrule  %
    \multicolumn{6}{l}{\textbf{Mistral Instruct 7B} (\texttt{mistral-7b-instruct-v0.2})}     \\
       \makecell[l]{zero-shot}          & 44.5$\pm$3.5                      & 53.0$\pm$5.0                     & 63.0$\pm$4.9                 & 38.0$\pm$3.4 & 46.8$\pm$2.0 & \textbf{6.02$\pm$0.04} & 18.50$\pm$0.66     \\
       \makecell[l]{RAG-cutoff}          & 48.0$\pm$3.5                       & 44.0$\pm$5.0                     & \textbf{71.0$\pm$4.6}                 & \underline{51.5$\pm$3.5} & 52.3$\pm$2.0 & \underline{5.90$\pm$0.05} & 17.82$\pm$0.68      \\
       \makecell[l]{narrative-experts \textbf{(Ours)}}          & \underline{55.0$\pm$3.5}                       & \underline{81.0$\pm$3.9}                     & 57.0$\pm$5.0                 & 42.5$\pm$3.5 & \underline{55.5$\pm$2.0} & 5.87$\pm$0.04 & \underline{20.57$\pm$0.71}      \\
       \makecell[l]{narrative-experts-RAG-cutoff \textbf{(Ours)}}          & \textbf{62.0$\pm$3.4}                       & \textbf{87.0$\pm$3.4}                     & \underline{66.0$\pm$4.8}                 & \textbf{58.5$\pm$3.5} & \textbf{65.7$\pm$1.9} & 5.85$\pm$0.04 & \textbf{22.20$\pm$0.80}      \\
       \midrule %
    \multicolumn{6}{l}{\textbf{GPT-3.5 Turbo} (\texttt{gpt-3.5-turbo-1106})}     \\
       \makecell[l]{zero-shot}          & 29.0$\pm$3.2                       & 33.0$\pm$4.7                     & \textbf{91.0$\pm$2.9}                 & 41.5$\pm$3.5 & 44.2$\pm$2.0 & \textbf{5.89$\pm$0.04} & 24.06$\pm$0.93      \\
       \makecell[l]{RAG-cutoff}          & 37.5$\pm$3.4                       & 34.0$\pm$4.8                     & \textbf{91.0$\pm$2.9}                 & \underline{55.5$\pm$3.5} & 51.8$\pm$2.0 & 5.73$\pm$0.05 & 24.39$\pm$0.95      \\
       \makecell[l]{narrative-experts \textbf{(Ours)}}          & \textbf{47.5$\pm$3.5}                       & \underline{70.0$\pm$4.6}                     & \underline{86.0$\pm$3.5}                 & 43.5$\pm$3.5 & \underline{56.3$\pm$2.0} & \underline{5.76$\pm$0.04} & \underline{27.03$\pm$0.92}      \\
       \makecell[l]{narrative-experts-RAG-cutoff \textbf{(Ours)}}          & \underline{46.0$\pm$3.5}                       & \textbf{72.0$\pm$4.5}                     & 84.0$\pm$3.7                 & \textbf{57.5$\pm$3.5} & \textbf{60.5$\pm$2.0} & 5.61$\pm$0.05 & \textbf{28.24$\pm$0.93}      \\
       \midrule  %
    \multicolumn{6}{l}{\textbf{GPT-4 Turbo} (\texttt{gpt-4-1106-preview})}     \\
        \makecell[l]{zero-shot}          & 46.5$\pm$3.5                       & 75.0$\pm$4.4                     & 90.0$\pm$3.0                 & 59.0$\pm$3.5 & 62.7$\pm$2.0 & 6.44$\pm$0.03 & 24.63$\pm$0.71      \\
        \makecell[l]{zero-shot-cot}          & 48.5$\pm$3.5                       & 75.0$\pm$4.4                     & \underline{92.0$\pm$2.7}                 & 61.0$\pm$3.5 & 64.3$\pm$2.0 & 6.51$\pm$0.03 & 23.67$\pm$0.65      \\
       \makecell[l]{few-shot}          & 47.0$\pm$3.5                       & 76.0$\pm$4.3                     & 88.0$\pm$3.3                 & 67.0$\pm$3.3 & 65.3$\pm$1.9 & 6.35$\pm$0.03 & 28.35$\pm$0.87     \\
       \makecell[l]{self-refine}          & 48.0$\pm$3.5                       & 75.0$\pm$4.4                     & \textbf{94.0$\pm$2.4}                 & 65.0$\pm$3.4 & 65.8$\pm$1.9 & 6.44$\pm$0.03 & 24.41$\pm$0.70      \\
       \makecell[l]{RAG}          & 33.5$\pm$3.4                       & 81.0$\pm$3.9                     & 91.0$\pm$2.9                 & \underline{72.0$\pm$3.2} & 63.8$\pm$2.0 & \textbf{6.55$\pm$0.02} & 21.14$\pm$0.64       \\
       \makecell[l]{RAG-cutoff}          & 50.0$\pm$3.5                       & 79.0$\pm$4.1                     & \underline{92.0$\pm$2.7}                 & \underline{72.0$\pm$3.2} & 69.2$\pm$1.9 & \underline{6.47$\pm$0.03} & 24.15$\pm$0.72       \\
       \makecell[l]{narrative-experts \textbf{(Ours)}}          & \underline{92.5$\pm$1.9}                       & \textbf{90.0$\pm$3.0}                     & 90.0$\pm$3.0                 & 67.5$\pm$3.3 & \underline{83.3$\pm$1.5} & 6.27$\pm$0.03 & \textbf{31.86$\pm$0.73}       \\
       \makecell[l]{narrative-experts-RAG-cutoff \textbf{(Ours)}}          & \textbf{93.0$\pm$1.8}                       & \underline{89.0$\pm$3.1}                     & 88.0$\pm$3.3                 & \textbf{74.5$\pm$3.1} & \textbf{85.3$\pm$1.5} & 6.30$\pm$0.03  & \underline{31.18$\pm$0.72}      \\
       \midrule  %
    \multicolumn{6}{l}{\textbf{GPT-4o} (\texttt{gpt-4o-2024-05-13})}     \\
      \makecell[l]{zero-shot}          & 46.0$\pm$3.5                       & 74.0$\pm$4.4                     & \underline{90.0$\pm$3.0}                 & 65.5$\pm$3.5 & 64.5$\pm$2.0 & \underline{6.26$\pm$0.03} & 26.78$\pm$0.81     \\
      \makecell[l]{RAG-cutoff}          & 51.0$\pm$3.5                       & 74.0$\pm$3.5                     & \textbf{92.0$\pm$2.7}                 & \underline{74.5$\pm$3.1} & 69.5$\pm$1.9 & \textbf{6.28$\pm$0.03} & 24.27$\pm$0.73     \\
      \makecell[l]{narrative-experts \textbf{(Ours)}}          & \underline{94.5$\pm$1.6}                       & \underline{84.0$\pm$3.7}                     & 83.0$\pm$3.8                 & 68.5$\pm$3.3 & \underline{82.2$\pm$1.6} & 6.02$\pm$0.04 & \textbf{33.58$\pm$0.80}     \\
      \makecell[l]{narrative-experts-RAG-cutoff \textbf{(Ours)}}          & \textbf{95.5$\pm$1.5}                       & \textbf{89.0$\pm$3.1}                     & 86.0$\pm$3.5                 & \textbf{79.5$\pm$2.9} & \textbf{87.5$\pm$1.4} & 6.05$\pm$0.04 & \underline{32.57$\pm$0.83}     \\
       \bottomrule
   \end{tabular}
   \end{adjustbox}
   \caption{Results of point-in-time character hallucination on 600 sampled data instances. We report the average scores with their standard error of the mean (SEM). \textbf{Bold} numbers indicate the highest scores, while \underline{underline} numbers are the second-best. All responses are evaluated by GPT-4 Turbo (\texttt{gpt-4-1106-preview}) as judges, with the exception of measuring AlignScore~\citep{Zha:2023:ACL}.}
   \label{tab:main_experiment}
\end{center}
\end{table*}

\subsection{Experimental Results}
See Appendix~\ref{subsec:implementation_details} for implementation details.
Table~\ref{tab:main_experiment} finds even GPT-4o and GPT-4, state-of-the-art LLMs, still struggle with point-in-time character hallucinations.

\textbf{\textit{Future} type.}
All baselines exhibit confusion with \textit{future} type questions with accuracies at  51\% or below.
It highlights a prevailing issue of role-playing agents that inadvertently disclose future events.
The naive RAG scores the lowest among baselines, showing that indiscriminately providing contexts harms the performance.
Our \methodName and \methodNameWithRag significantly enhance performance, thanks to the temporal expert.

\textbf{\textit{Past-absence} and \textit{past-only} types.}
Both naive RAG and RAG-cutoff can potentially mitigate hallucinations for these question types by leveraging context from their retrieval modules.
However, their performance still lags behind that observed in \textit{past-presence} questions, with gaps of 10\% points and 13\% points in \textit{past-absence} types, and 19\% points and 20\% points in \textit{past-only} types, respectively.
Conversely, our methods enhance outcomes in both \textit{past-absence} and \textit{past-only} types, thanks to the support of both temporal and spatial experts.

\textbf{\textit{Past-presence} type instances:}
All baselines, except for Mistral, perform admirably, showcasing the role-playing LLMs' proficiency in memorizing narratives from novel series.
Our methods slightly lag in this type due to narrative experts' occasional mispredictions, yet this shortfall is minor compared to significant enhancement in the other three types.

\textbf{Personality consistency:}
All methods generally maintain a consistent character portrayal, scoring above 5.6 in personality consistency.
However, our methods receive lower scores from the GPT-4 judge due to their tendency to respond with unawareness regarding future events or character absences, which sometimes falls short of fully conveying the expected character's personality.
In contrast, the GPT-4 judge appears to favor responses from role-playing agents that indiscriminately disclose information, regardless of its relevance to the character's knowledge boundary at a specific time point.

\textbf{AlignScore evaluation:}
In addition, we utilize AlignScore~\citep{Zha:2023:ACL}, a metric based on a post-trained RoBERTa-large model, to assess the spatiotemporal consistency of the role-playing agent without relying on GPT-4 judges.
Refer to Appendix~\ref{subsec:experimental_result_11k} for details of the AlignScore.
The results in Table~\ref{tab:main_experiment} show that AlignScore is in agreement with evaluations from GPT-4 judges, and our methods achieve the highest AlignScores across the three different backbone LLMs.

Furthermore, we provide further analyses beyond the main experiments, including human evaluation results that closely align with those of the LLM judges, as detailed in Appendix~\ref{sec:further_analyses}.

\section{Conclusion}
We highlighted the importance of point-in-time role-playing agents for enhancing narrative engagement, preventing spoilers, and facilitating fandom role-play activities.
To maintain a character's spatiotemporal consistency and avoid hallucinations, we introduced the \datasetName\ benchmark and developed an automated pipeline, resulting in 10,895 instances.
Using \datasetName, we identified significant hallucination issues in state-of-the-art role-playing LLMs.
To address these, we proposed \methodName, an effective method to reduce character hallucinations by breaking down the reasoning process and guiding it with narrative experts.
Despite these efforts, our findings indicate ongoing challenges with point-in-time character hallucinations, suggesting the need for further improvements.

\section*{Limitations}
\label{sec:limitation}

Despite the advancements presented in this study, there are some limitations as follows.
(1) Sourced only from English books: Since \datasetName consists primarily of texts written in English and sourced from English-speaking countries, it may reflect cultural biases inherent to these regions. One solution is to incorporate multilingual and multicultural books, as new data can be added to \datasetName automatically.
(2) High costs of GPT-4 judges: The financial expenses of extensive GPT-4 evaluations can be prohibitive, restricting the feasibility of conducting large-scale assessments. An alternative would be using open-source LLMs for evaluation, such as~\citet{Kim:2024:ICLR}.
(3) Latency and cost issues with \methodName: The narrative expert requires generating multiple hints and responses per question. This introduces increased latency and computational costs. Future research on efficiently generating responses while reducing point-in-time character hallucinations will be anticipated.

\section*{Ethics Statement}\label{sec:ethical}
To mitigate any potential issue arising from the use of the four novel series, we address concerns about copyright issues as follows:

\begin{enumerate}
    \item Source attribution: The dataset utilizes raw text from each novel series, and we acknowledge the copyrights held by the authors and publishers.
    \item Fair use justification: We believe our use of the copyrighted text qualifies as ``fair use'' under U.S. law, whose criteria include:
    \begin{itemize}
        \item Purpose of use: The dataset is used exclusively for non-commercial, educational, and research purposes.
        \item Nature of the copyrighted work: The work is used in a research context to evaluate point-in-time character hallucination of role-playing LLMs, an inherently academic pursuit.
        \item Lack of market harm: Our dataset does not substitute for the original works nor harm their market.
    \end{itemize}
    \item Content of dataset: \datasetName comprises only a fraction of the content necessary for dataset construction, indicating that our dataset includes approximately 40\% of the original text from the sources.
    \item Accessibility and reproducibility: We will publish all dataset scripts and the dataset itself, restricting access to those who agree to use it only for research.
\end{enumerate}

Besides, the generated dataset may inadvertently include harmful content intended to mislead characters.
In adherence to the NLP ethics community’s guidelines on `toxic text'~\citep{Gehman:2020:EMNLPF,Askell:2021:arxiv}, We manually reviewed all 11K data instances and filtered out those containing provocative scenes.
Specifically, we removed fewer than ten cases associated with severe trauma and explicit violence that could negatively impact users who will read questions and the agent’s responses.

By implementing these measures, we ensure that our research respects both the legal rights of the original content creators and the ethical standards of the research community.

\section*{Acknowledgements}
We thank Jamin Shin, Hyunwoo Kim, Euihyun Tae, and the anonymous reviewers for their valuable comments.
This work was supported by SNU-NAVER Hyperscale AI Center, the Institute of Information \& Communications Technology Planning \& Evaluation (IITP) grant funded by the Korea government (MSIT) (No.~RS-2019-II191082, SW StarLab; No.~RS-2022-II220156, Fundamental research on continual meta-learning for quality enhancement of casual videos and their 3D metaverse transformation; No.~RS-2021-II211343, Artificial Intelligence Graduate School Program (Seoul National University)), and the National Research Foundation of Korea (NRF) grant funded by the Korea government (MSIT) (No.~2023R1A2C2005573).
Gunhee Kim is the corresponding author.


\bibliography{acl2024}

\appendix

\newpage

{\renewcommand{\arraystretch}{1.0}
    \begin{table*}[t!] \begin{center}
    \begin{adjustbox}{width=1.0\textwidth}
    \begin{tabular}{lcccccc}
        \toprule
        \makecell[l]{Evaluation\\Dataset /\\Benchmark} & \makecell{Dataset\\automatically\\constructed?} & \makecell{Support\\point-in-time\\role-playing?} & \makecell{Evaluate\\near-future\\unawareness?\\(Temporal)} & \makecell{Evaluate\\absence\\awareness?\\(Spatial)} & \makecell{Evaluate\\fake event\\awareness?\\(Fake question)} & \makecell{Evaluation method\\for character\\hallucination} \\
        \midrule
        \multicolumn{7}{l}{\textbf{\textit{Role-Playing Domain} (Concurrent)}}     \\
        \makecell[l]{RoleEval \small (2023/12)} & \xmark & \xmark & \xmark$^*$ & $\Delta$ & \xmark & \makecell{\small Answer selection accuracy:\\\small $[$0 or 1$]$ (selection-based)} \\
        \makecell[l]{SimulateBench \small (2023/12)} & \xmark & \xmark & \xmark & \xmark & \xmark & \makecell{\small Answer selection accuracy:\\\small $[$0 or 1$]$ (selection-based)} \\
        \makecell[l]{RoleInstruct \small (2024/01)} & \cmark & \xmark & \xmark & \xmark & \xmark & \makecell{\small Human as judges:\\\small $[$1 to 5$]$ (unscalable)} \\
        \makecell[l]{CharacterEval \small (2024/01)} & \cmark & \xmark & \xmark & \xmark & \xmark & \makecell{\small CharacterRM:\\\small $[$1 to 5$]$ (trainable)} \\
        \makecell[l]{WikiRole \small (2024/01)} & \cmark & \xmark & \makecell{\xmark \small (Question from\\\small distinct era/narrative:\\\small easy)} & \xmark & \xmark & \makecell{\small LLM as judges w/\\\small temporal labels:\\\small $[$0 or 1$]$ (accurate)} \\
        \makecell[l]{MORTISE \small (2024/02)} & \cmark & \xmark & \xmark & \xmark & \cmark & \makecell{\small RC.Score:\\\small $[$1 to 5$]$ (trainable)} \\
        \makecell[l]{Character100 \small (2024/03)} & \cmark & \xmark & \xmark & \xmark & \xmark & \makecell{\small Rouge-L w/ gold response:\\\small $[$0-1$]$ (implicit)} \\
        \makecell[l]{RoleInteract \small (2024/03)} & \cmark & \xmark & \xmark & \xmark & \xmark & \makecell{\small Answer selection accuracy:\\\small $[$0 or 1$]$ (selection-based)} \\
        \midrule
        \makecell[l]{\textbf{\datasetName} \small (2024/02 submitted)} & \cmark & \cmark & \makecell{\cmark \small (Question from\\\small the same era/narrative:\\\small hard)} & \cmark & \cmark & \makecell{\small LLM as judges w/\\\small spatiotemporal labels:\\\small $[$0 or 1$]$ (accurate)} \\
        \bottomrule
    \end{tabular}
    \end{adjustbox}
    \caption{Comparison of \datasetName with concurrent role-playing datasets or benchmarks released in or after December 2023: RoleEval~\citep{Shen:2023:arxiv}, SimulateBench~\citep{Xiao:2023:arxiv}, RoleInstruct~\citep{Tao:2024:arxiv}, CharacterEval~\citep{Tu:2024:arxiv}, WikiRole~\citep{Lu:2024:arxiv}, MORTISE~\citep{Tang:2024:arxiv}, Character100~\citep{Wang:2024:LREC}, and RoleInteract~\citep{Chen:2024:arxiv}. Note that RoleEval evaluates whether LLMs possess specific role knowledge by having them answer multiple-choice questions, rather than evaluating role-playing LLM agents. *It supports evaluating timeline reasoning, where LLMs are required to sort related events in temporal order. However, this does not mean it evaluates \textit{near-future unawareness} because it assumes that LLMs are omniscient about the characters. $\Delta$ indicates that RoleEval does not explicitly evaluate \textit{absence awareness} since it lacks role-playing scenarios, but it implicitly supports reasoning about event participants (\ie identifying who took part in an event). In the last column, `selection-based' means that the task involves selecting the correct answer for multiple-choice questions rather than generating an open-ended response. `Trainable' means that the reward models, CharacterRM and RC.Score, are trained using datasets evaluated by human judges. Although those models are more aligned with human judges than GPT-4 judges, it is unknown whether they are still effective at evaluating LLMs that role-play out-of-domain characters, characters not in the training dataset.}
    \label{tab:concurrent_dataset_comparison}
\end{center}\end{table*}}

\section{Related Work (Full ver.)}
\label{sec:updated_related_work}
\paragraph{Concurrent role-play benchmarks.}
We compare \datasetName to concurrent role-playing datasets or benchmarks~\citep{Shen:2023:arxiv,Xiao:2023:arxiv,Tao:2024:arxiv,Tu:2024:arxiv,Lu:2024:arxiv,Tang:2024:arxiv,Wang:2024:LREC,Chen:2024:arxiv} released in or after December 2023, as detailed in Table~\ref{tab:concurrent_dataset_comparison}.
Note that RoleEval~\citep{Shen:2023:arxiv} assesses whether LLMs possess specific role knowledge by having them answer multiple-choice questions, rather than evaluating role-playing LLM agents.
While some concurrent benchmarks partially address the goals of \datasetName intended to do (\eg evaluation of event participant reasoning in RoleEval~\citep{Shen:2023:arxiv}, evaluation of fake event awareness in MORTISE~\citep{Tang:2024:arxiv}, use of temporal labels for LLM judges in WikiRole~\citep{Lu:2024:arxiv}), \datasetName remains the most comprehensive benchmark for evaluating point-in-time character hallucination.

\paragraph{LLM hallucinations.}
Since LLMs are prone to hallucinations~\citep{Ji:2023:ACMCS,Zhang:2023:arxiv}, many studies have evaluated hallucinations in generated texts~\citep{Zha:2023:ACL,Min:2023:EMNLP,Mishra:2024:arxiv,Hong:2024:arxiv}.
To reduce hallucination, some works have focused on retrieval-augmented generation~\citep{Lewis:2020:NeurIPS, Shuster:2021:EMNLPF, Mialon:2023:TMLR}, while others have incorporated honesty alignment (\ie refusing to answer questions when LLMs lack knowledge) in text generation~\citep{Yang:2023:arxiv, Cheng:2024:arxiv}.

Unlike prior works that address general hallucination issues, we specifically focus on \textit{point-in-time character} hallucination, which includes unique spatiotemporal consistency problems such as \textit{future unawareness}, \textit{absence awareness}, and \textit{fake event awareness}. Furthermore, we propose a new method, \methodName, to mitigate these hallucinations.

\paragraph{Neural theory-of-mind in narrative understanding.}
Neural theory-of-mind (ToM) examines whether LLMs possess the ability to understand the mental stages (\eg thoughts, beliefs, and intentions) of others~\citep{Sap:2022:EMNLP,Kim:2023:EMNLP}.
In narrative understanding, ToM capabilities are essential for role-playing agents to comprehend both the narratives and the characters' minds~\citep{Sang:2022:NAACL,Yu:2022:arxiv,Zhao:2024:arxiv}.

Although it is well known that LLMs memorize extensive knowledge from books~\citep{Chang:2023:EMNLP} and can precisely answer questions about narrative (See Table~\ref{tab:narrative_experts_accuracy} in Appendix~\ref{subsec:narrative_experts_accuracy}), we found that they cannot maintain spatiotemporal consistency while acting as point-in-time role-playing agents, as shown in Table~\ref{tab:main_experiment}.
The results demonstrate that LLMs are not yet capable of ToM in point-in-time role-playing scenarios.

\section{Prompt Demonstration}
\label{sec:prompt_demonstration}
\paragraph{Prompts for GPT-4 Turbo judges.}
We present the prompts used for evaluating the two dimensions as follows:
\begin{enumerate}
    \item Spatiotemporal consistency: refer to Table~\ref{tab:spatiotemporal_consistency_prompt}.
    \item Personality consistency: refer to Table~\ref{tab:personality_consistency_prompt}.
\end{enumerate}

\begin{table*}[htbp]
\scriptsize
\centering
\begin{tabular}{@{}p{\linewidth}@{}}
\toprule
\textbf{Prompt for Spatiotemporal Consistency Evaluation}\\
\midrule
You will be given responses written by an AI assistant mimicking the character \{agent\_name\}. Your task is to rate the performance of \{agent\_name\} using the specific criterion by following the evaluation steps. Below is the data:

~

***

[Interactions]

Interviewer: \{question\}

\{agent\_name\}: \{response\}

***

[Fact]

\{spatiotemporal\_label\}

~

[Evaluation Criterion]

Spatiotemporal Consistency (0 or 1): Is the response consistent with the character's 
spatiotemporal knowledge?

~

[Evaluation Steps]

1. Read through the [Fact] and identify the knowledge scope of the character.

2. Read through the interactions and responses of the AI assistant to find the evidence of knowledge used in the response.

3. Compare the evidence to the [Fact]. Check if the response is consistent with the character's knowledge scope.

4. If some knowledge contradicts or contains inconsistencies about the [Fact], given a 0 score. Otherwise, assign a 1 score.

***

~

First, write out in a step by step manner your reasoning about the criterion to be sure that your conclusion is correct. Avoid simply stating the correct answers at the outset. Then, print the score on its own line corresponding to the correct answer. At the end, repeat just the selected score again by itself on a new line. \\
\bottomrule
\end{tabular}
    \caption{Prompt for GPT-4 Turbo judges to evaluate spatiotemporal consistency.}
    \label{tab:spatiotemporal_consistency_prompt}
\end{table*}

\begin{table*}[htbp]
\scriptsize
\centering
\begin{tabular}{@{}p{\linewidth}@{}}
\toprule
\textbf{Prompt for Personality Consistency Evaluation}\\
\midrule
You will be given responses written by an AI assistant mimicking the character \{agent\_name\}. Your task is to rate the performance of \{agent\_name\} using the specific criterion by following the evaluation steps. Below is the data:

~

***

[Interactions]

Interviewer: \{question\}

\{agent\_name\}: \{response\}

***

[Personality]

\{personality\_label\}

~

[Evaluation Criterion]

Personality Consistency (1-7): Is the response consistent with the character's personality?

~

[Evaluation Steps]

1. Read through the [Personality] and write the personalities, including preferences, values, and convictions of the real character.

2. Read through the interactions and identify the personalities, including preferences, values, and convictions of the AI assistant.

3. After having a clear understanding of the interactions, compare the response to the [Personality]. Look for any consistencies or inconsistencies. Do the responses reflect the character’s personalities, including preferences, values, and convictions?

4. Use the given scale from 1-7 to rate how well the response reflects the personalities, including preferences, values, and convictions of the character. 1 being not at all reflective of the character’s personalities, and 7 being perfectly reflective of the character’s personalities.

***

~

First, write out in a step by step manner your reasoning about the criterion to be sure that your conclusion is correct. Avoid simply stating the correct answers at the outset. Then, print the score on its own line corresponding to the correct answer. At the end, repeat just the selected score again by itself on a new line. \\
\bottomrule
\end{tabular}
    \caption{Prompt for GPT-4 Turbo judges to evaluate personality consistency.}
    \label{tab:personality_consistency_prompt}
\end{table*}

\section{Examples of \datasetName}
\label{sec:examples_of_timechara}
We show examples of our dataset across four different data types: \textit{future}, \textit{past-absence}, \textit{past-presence}, and \textit{past-only}, each illustrated as follows:
\begin{enumerate}
    \item \textit{future}: An instance of \textit{future} data is presented in Table~\ref{tab:fact_structured_future_example}, characterized by its \textit{fact-based structured} question.
    \item \textit{past-absence}: Table~\ref{tab:fact_structured_past_absence_example} illustrates the \textit{past-absence} type with a similarly \textit{fact-based structured} question.
    \item \textit{past-presence}: Table~\ref{tab:fact_structured_past_presence_example} shows an example of \textit{past-presence} type data, whose question is also a \textit{fact-based structured} question.
    \item \textit{past-only}: Demonstrations of \textit{past-only} type data are provided in Table~\ref{tab:fact_freeform_past_only_example} for a \textit{fact-based free-form} question and in Table~\ref{tab:fake_freeform_past_only_example} for a \textit{fake-based free-form} question.
\end{enumerate}

\begin{table*}[t!] \begin{center}
    \begin{adjustbox}{width=\linewidth}
   \begin{tabular}{ll}
       \toprule
       \makecell[l]{\textbf{Scene}} & \makecell[l]{``Why can't we get through?'' Harry hissed to Ron... ``I think we'd better go and wait by the car,'' said Harry...} \\
       \midrule
       \makecell[l]{\textbf{Event Summary}} & \makecell[l]{Harry and Ron took the enchanted car to Hogwarts after a barrier mishap at King’s Cross.} \\
       \midrule
       \makecell[l]{\textbf{Question}} & \makecell[l]{``Tell me your feelings when \{Event Summary\}.''} \\
       \midrule
       \makecell[l]{\textbf{Character}} & \makecell[l]{2nd-year Hermione Granger on Halloween} \\
       \midrule
       \makecell[l]{\textbf{Data Type}} & \makecell[l]{Past-absence} \\
       \midrule
       \makecell[l]{\textbf{Spatiotemporal Label}} & \makecell[l]{\textbf{Past}: During her 2nd-year on Christmas, Hermione Granger can respond based on the moment but should not\\wrongly recall it.\\- Moment: \{Scene\}\\\textbf{Absence}: During her 2nd-year on Christmas, Hermione Granger should not say that she was present when \{Event Summary\}.}\\
       \midrule
       \makecell[l]{\textbf{Personality Label}} & \makecell[l]{Hermione Granger is portrayed as exceptionally intelligent and hardworking, often topping her classes and ...} \\
       \midrule
       \makecell[l]{\textbf{Gold Response}} & \makecell[l]{``Oh no, I wasn't there when it happened. I'd already gone through the barrier and was on the Hogwarts Express...''} \\
       \bottomrule
   \end{tabular}
   \end{adjustbox}
   \caption{An example of our \textit{past-absence} type data instance with the \textit{fact-based structured} question.}
   \label{tab:fact_structured_past_absence_example}
\end{center}
\end{table*}

\begin{table*}[t!] \begin{center}
    \begin{adjustbox}{width=\linewidth}
   \begin{tabular}{ll}
       \toprule
       \makecell[l]{\textbf{Scene}} & \makecell[l]{``Why can't we get through?'' Harry hissed to Ron... ``I think we'd better go and wait by the car,'' said Harry...} \\
       \midrule
       \makecell[l]{\textbf{Event Summary}} & \makecell[l]{Harry and Ron took the enchanted car to Hogwarts after a barrier mishap at King’s Cross.} \\
       \midrule
       \makecell[l]{\textbf{Question}} & \makecell[l]{``Tell me your feelings when \{Event Summary\}.''} \\
       \midrule
       \makecell[l]{\textbf{Character}} & \makecell[l]{2nd-year Ronald Weasley on Halloween} \\
       \midrule
       \makecell[l]{\textbf{Data Type}} & \makecell[l]{Past-presence} \\
       \midrule
       \makecell[l]{\textbf{Spatiotemporal Label}} & \makecell[l]{\textbf{Past}: During his 2nd-year on Halloween, Ronald Weasley can respond based on the moment but should not\\wrongly recall it.\\- Moment: \{Scene\}.\\\textbf{Presence}: During his 2nd-year on Halloween, Ronald Weasley should not say that he was absent when \{Event Summary\}.} \\
       \midrule
       \makecell[l]{\textbf{Personality Label}} & \makecell[l]{Ronald Weasley is depicted as a loyal, brave, strong, and humorous individual, yet sometimes immature and ...} \\
       \midrule
       \makecell[l]{\textbf{Gold Response}} & \makecell[l]{``Blimey, yeah, I was there, wasn't I? It was mental. One minute we're running towards the barrier between...''} \\
       \bottomrule
   \end{tabular}
   \end{adjustbox}
   \caption{An example of our \textit{past-presence} type data instance with the \textit{fact-based structured} question.}
   \label{tab:fact_structured_past_presence_example}
\end{center}
\end{table*}

\begin{table*}[t!] \begin{center}
    \begin{adjustbox}{width=\linewidth}
   \begin{tabular}{ll}
       \toprule
       \makecell[l]{\textbf{Scene}} & \makecell[l]{``Why can't we get through?'' Harry hissed to Ron... ``I think we'd better go and wait by the car,'' said Harry...} \\
       \midrule
       \makecell[l]{\textbf{Event Summary}} & \makecell[l]{Harry and Ron took the enchanted car to Hogwarts after a barrier mishap at King’s Cross.} \\
       \midrule
       \makecell[l]{\textbf{Question}} & \makecell[l]{``Why did Harry and Ron use the enchanted car to get to Hogwarts?''} \\
       \midrule
       \makecell[l]{\textbf{Character}} & \makecell[l]{2nd-year Harry Potter at the end of the scene} \\
       \midrule
       \makecell[l]{\textbf{Data Type}} & \makecell[l]{Past-only} \\
       \midrule
       \makecell[l]{\textbf{Spatiotemporal Label}} & \makecell[l]{\textbf{Past}: At the end of the scene of Harry Potter and the Chamber of Secrets as a 2nd-year student,\\Harry Potter can respond based on the moment but should not wrongly recall it.\\- Moment: \{Scene\}.\\- Answer: ... Due to crashing into the barrier at King's Cross Station, prompting them to fly to Hogwarts...} \\
       \midrule
       \makecell[l]{\textbf{Personality Label}} & \makecell[l]{Harry Potter is characterized by his selflessness and immense loyalty, especially towards his friends...} \\
       \midrule
       \makecell[l]{\textbf{Gold Response}} & \makecell[l]{``Well, it was a bit of a mess, really. Ron and I tried to go through the barrier to Platform Nine and Three-Quarters...''} \\
       \bottomrule
   \end{tabular}
   \end{adjustbox}
   \caption{An example of our \textit{past-only} type data instance with the \textit{fact-based free-form} question.}
   \label{tab:fact_freeform_past_only_example}
\end{center}
\end{table*}

\begin{table*}[t!] \begin{center}
    \begin{adjustbox}{width=\linewidth}
   \begin{tabular}{ll}
       \toprule
       \makecell[l]{\textbf{Scene}} & \makecell[l]{``Why can't we get through?'' Harry hissed to Ron... ``I think we'd better go and wait by the car,'' said Harry...} \\
       \midrule
       \makecell[l]{\textbf{Event Summary}} & \makecell[l]{Harry and Ron took the enchanted car to Hogwarts after a barrier mishap at King’s Cross.} \\
       \midrule
       \makecell[l]{\textbf{Question}} & \makecell[l]{``Why did Harry and Ron consider swimming to Hogwarts after the barrier incident at King’s Cross?''} \\
       \midrule
       \makecell[l]{\textbf{Character}} & \makecell[l]{2nd-year Harry Potter at the end of the scene} \\
       \midrule
       \makecell[l]{\textbf{Data Type}} & \makecell[l]{Past-only} \\
       \midrule
       \makecell[l]{\textbf{Spatiotemporal Label}} & \makecell[l]{\textbf{Past}: At the end of the scene of Harry Potter and the Chamber of Secrets as a 2nd-year student,\\Harry Potter can respond based on the moment but should not wrongly recall it.\\- Moment: \{Scene\}.\\- Answer: Harry and Ron did not decide to swim to Hogwarts; they decided to fly to Hogwarts in the Weasley's...} \\
       \midrule
       \makecell[l]{\textbf{Personality Label}} & \makecell[l]{Harry Potter is characterized by his selflessness and immense loyalty, especially towards his friends...} \\
       \midrule
       \makecell[l]{\textbf{Gold Response}} & \makecell[l]{``Oh, blimey, swim to Hogwarts? That would've been a right mess, wouldn't it? No, no, we didn't swim...''} \\
       \bottomrule
   \end{tabular}
   \end{adjustbox}
   \caption{An example of our \textit{past-only} type data instance with the \textit{fake-based free-form} question.}
   \label{tab:fake_freeform_past_only_example}
\end{center}
\end{table*}

\begin{table*}[htbp]
\scriptsize
\centering
\begin{tabular}{@{}p{\linewidth}@{}}
\toprule
\textbf{Prompt for Scene and Speakers List Extraction}\\
\midrule

First, read chapter 1, part 1 of the book ``The Lord of the Rings - The Fellowship of the Ring''. Then, extract 5 parts from the raw text with dialogues that can be considered as one scene. Each part should meet the following requirements. Start by analyzing the text that I gave you.

~

1. Each scene should be unique throughout the entire series: The Lord of The Ring.

2. Each scene shouldn't be ambiguous, which means that the summary of each scene should be talking or related to a specific event, item, or person.

3. Scenes shouldn't be everyday conversation such as the summary of the scene being: ``Frodo Baggins talked to Sam about his breakfast'', which could be an everyday conversation.

4. Each scene should contain at least 5 dialogues. The extracted raw text should be between 15 to 35 sentences long to sufficiently form the scene and contain sufficient information about the scene.

~

For each scene, please provide:

1. A summary of the scene you selected in one sentence.

2. The raw text that you selected.

3. The full name of the characters speaking in that scene.

***

Input:

- Raw Text: \{raw\_text\}

~

Output: \\
\bottomrule
\end{tabular}
    \caption{An example of a prompt for extracting scenes and speakers list from the Lord of the Rings series.}
    \label{tab:data_prompt_scene_speakers_extraction}
\end{table*}
\section{Details of Dataset Construction}
\label{sec:dataset_construction_details}
\subsection{Extract scenes, event summaries, and participant lists from book}
\label{subsec:scene_extraction}
For the Harry Potter series, we utilize the scene dataset from \citet{Chen:2023:EMNLPF}, which comprises manually selected 1,037 unique scenes from the books.
For the other three series, we extract 300 distinct scenes containing multi-turn dialogues among characters, as detailed in Table~\ref{tab:data_prompt_scene_speakers_extraction}.
Note that we include multi-turn dialogues among characters to ensure that each scene features interaction among participants, with at least one character involved.
In addition, Table~\ref{tab:data_prompt_summary_participants_generation} shows an example of a prompt for generating an event summary and participant lists.

\begin{table*}[htbp]
\scriptsize
\centering
\begin{tabular}{@{}p{\linewidth}@{}}
\toprule
\textbf{Prompt for Event Summary and Participant List Generation}\\
\midrule

First, read the scene and dialogue. Then, generate a single, unique ``fact'' sentence in the past tense that captures Harry's distinct moment or experience that is retrievable from the scene. If there are several moments, pick the most unique moment and write it. Write it concisely. Finally, extract participants who are physically present and existing in the scene.

***

Input:

- Position: Book1-chapter2

- Speakers: Petunia, Vernon, Harry

- Scene:
“Bad news, Vernon,” she said. “Mrs. Figg’s broken her leg. She can’t take him.” She jerked her head in Harry’s direction.
Dudley’s mouth fell open in horror, but Harry’s heart gave a leap. Every year on Dudley’s birthday, his parents took him and a friend out for the day, to adventure parks, hamburger restaurants, or the movies. Every year, Harry was left behind with Mrs. Figg, a mad old lady who lived two streets away. Harry hated it there. The whole house smelled of cabbage and Mrs. Figg made him look at photographs of all the cats she’d ever owned.
“Now what?” said Aunt Petunia, looking furiously at Harry as though he’d planned this. Harry knew he ought to feel sorry that Mrs. Figg had broken her leg, but it wasn’t easy when he reminded himself it would be a whole year before he had to look at Tibbles, Snowy, Mr. Paws, and Tufty again.
“We could phone Marge,” Uncle Vernon suggested.
“Don’t be silly, Vernon, she hates the boy.”
The Dursleys often spoke about Harry like this, as though he wasn’t there — or rather, as though he was something very nasty that couldn’t understand them, like a slug.
“What about what’s-her-name, your friend — Yvonne?”
“On vacation in Majorca,” snapped Aunt Petunia.
“You could just leave me here,” Harry put in hopefully (he’d be able to watch what he wanted on television for a change and maybe even have a go on Dudley’s computer).
Aunt Petunia looked as though she’d just swallowed a lemon.
“And come back and find the house in ruins?” she snarled.
“I won’t blow up the house,” said Harry, but they weren’t listening.
“I suppose we could take him to the zoo,” said Aunt Petunia slowly, “. . . and leave him in the car. . . .”
“That car’s new, he’s not sitting in it alone. . . .”
Dudley began to cry loudly. In fact, he wasn’t really crying — it had been years since he’d really cried — but he knew that if he screwed up his face and wailed, his mother would give him anything he wanted.
“Dinky Duddydums, don’t cry, Mummy won’t let him spoil your special day!” she cried, flinging her arms around him.
“I . . . don’t . . . want . . . him . . . t-t-to come!” Dudley yelled between huge, pretend sobs. “He always sp-spoils everything!” He shot Harry a nasty grin through the gap in his mother’s arms.
Just then, the doorbell rang —“Oh, good Lord, they’re here!” said Aunt Petunia frantically — and a moment later, Dudley’s best friend, Piers Polkiss, walked in with his mother. Piers was a scrawny boy with a face like a rat. He was usually the one who held people’s arms behind their backs while Dudley hit them. Dudley stopped pretending to cry at once.
Half an hour later, Harry, who couldn’t believe his luck, was sitting in the back of the Dursleys’ car with Piers and Dudley, on the way to the zoo for the first time in his life. His aunt and uncle hadn’t been able to think of anything else to do with him, but before they’d left, Uncle Vernon had taken Harry aside.
“I’m warning you,” he had said, putting his large purple face right up close to Harry’s, “I’m warning you now, boy — any funny business, anything at all — and you’ll be in that cupboard from now until Christmas.”

~

Output:

- Unique Fact: The Dursleys reluctantly decided to take Harry to the zoo with them for the first time in his life but warned him of severe consequences if he caused any trouble.

- Participants: Aunt Petunia, Dudley Dursley, Harry Potter, Uncle Vernon Dursley, Mrs. Figg, Piers Polkiss

***

Input:

- Position: \{position\}

- Speakers: \{speakers\}

- Scene:
\{extracted\_scene\}

~

Output: \\
\bottomrule
\end{tabular}
    \caption{An example of a prompt for generating an event summary and participant list for the Harry Potter series.}
    \label{tab:data_prompt_summary_participants_generation}
\end{table*}

\begin{table*}[htbp]
\scriptsize
\centering
\begin{tabular}{@{}p{\linewidth}@{}}
\toprule
\textbf{Prompt for Fact-based Free-from Question Generation}\\
\midrule

First, read the event summary from the Harry Potter series. Then, paraphrase the event summary to (1) a single-sentence question among 5w1h questions and (2) the answer to the question that should be answerable from the given event summary. Don't use pronouns to indicate the event, but self-contain what event it is. Note that the question should identify the unique period of the story.

***

Input:

- Event summary: Ron's broken wand caused the charm to backfire, erasing Lockhart's memory and causing a portion of the ceiling to cave in.

~

Output:

- Question: What caused Gilderoy Lockhart's memory loss and the partial collapse of the ceiling?

- Answer: Gilderoy Lockhart's memory was erased, and a portion of the ceiling caved in when Ron Weasley's broken wand caused a backfired charm in their second year at Hogwarts.

***

Input:

- Event summary: Harry uncovered that it was Professor Quirrell who attempted to seize the Sorcerer's Stone, revealing that he was under the influence of Lord Voldemort, who existed parasitically on the reverse side of Quirrell's head.

~

Output:

- Question: Who did Harry Potter find out was attempting to steal the Sorcerer's Stone and was possessed by Lord Voldemort during their encounter at Hogwarts, and where was Voldemort residing on the individual's body?

- Answer: Harry Potter discovered that Professor Quirrell, with Lord Voldemort residing on the back of his head, was trying to steal the Sorcerer's Stone.

***

Input:

- Event summary: \{event\_summary\}

~

Output: \\
\bottomrule
\end{tabular}
    \caption{An example of a prompt for generating a \textit{fact-based free-form} question for the Harry Potter series.}
    \label{tab:data_prompt_freeform_question_generation}
\end{table*}

\begin{table*}[htbp]
\scriptsize
\centering
\begin{tabular}{@{}p{\linewidth}@{}}
\toprule
\textbf{Prompt for Fake Event Summary Generation}\\
\midrule

First, read the event summary from the Harry Potter series. Generate the fake event summary that converts the true event summary to confuse readers using one of the six methods as follows.

~

1. Change the character: Swap the character with another character.

- True: Harry tricked Malfoy into freeing Dobby by giving Malfoy one of his own socks, which he promptly threw away and was caught by Dobby.

- Fake: Harry tricked Snape into freeing Dobby by giving Snape one of his own socks, which he promptly threw away and was caught by Dobby.

~

2. Change the Key Object: Alter the object that is central to the event.

- True: Harry used his own sock to free Dobby.

- Fake: Harry used a spellbook to free Dobby.

~

3. Alter the Location: Change the setting where the event took place.

- True: The event took place in Malfoy Manor.

- Fake: The event took place in the Gryffindor common room.

~

4. Switch the Action: Change what was done to the object or the action taken by the character.

- True: Malfoy threw the sock away.

- Fake: Malfoy donated the sock to charity.

~

5. Introduce a Nonexistent Character or Object: Add someone or something that wasn't originally there.

- True: Harry and Malfoy were the main characters involved.

- Fake: Harry, Malfoy, and a ghost named Sir Pudding were involved in the exchange.

~

6. Change the Character’s Knowledge: Switch what the character knows or doesn't know.

- True: Harry knew the sock would free Dobby.

- Fake: Harry had no idea that the sock would free Dobby and thought it was just a useless gift.

***

Input:

- True event summary: Harry received a Nimbus 2000, a gift from Professor McGonagall.

~

Output:

- Fake event summary: Harry received a Nimbus 2000, a gift from Professor Snape.

- Method number: 1. Change the character

***

Input:

- True event summary: Fred, George, and Ron rescued Harry from the Dursleys with the use of a Flying Ford Anglia.

~

Output:

- Fake event summary: Fred, George, and Ron rescued Harry from Hogwarts with the use of a Flying Ford Anglia.

- Method number: 3. Alter the Location

***

Input:

- True event summary: \{true\_event\_summary\}

~

Output: \\
\bottomrule
\end{tabular}
    \caption{An example of a prompt for generating a fake event summary for the Harry Potter series.}
    \label{tab:data_prompt_fake_summary_generation}
\end{table*}

\begin{table*}[htbp]
\scriptsize
\centering
\begin{tabular}{@{}p{\linewidth}@{}}
\toprule
\textbf{Prompt for Fake-based Free-from Question Generation}\\
\midrule

First, read two event summaries from the Harry Potter Series. One is a true event summary, and the other is a fake event summary, which is generated from the true event summary to confuse readers. Then, paraphrase the fake event summary to (1) a single-sentence fake question among 5w1h questions and (2) the true answer to the question that should be answerable from the given true event summary. Don't use pronouns to indicate the event, but self-contain what event it is. Note that the question should identify the unique period of the story.

***

Input:

- True event summary: Harry received a Nimbus 2000, a gift from Professor McGonagall.

- Fake event summary: Harry received a Nimbus 2000, a gift from Professor Snape.

~

Output:

- Fake question: Why did Professor Snape give Harry a Nimbus 2000?

- True answer: Professor Snape did not give Harry a Nimbus 2000; it was a gift from Professor McGonagall.

***

Input:

- True event summary: Fred, George, and Ron rescued Harry from the Dursleys with the use of a Flying Ford Anglia.

- Fake event summary: Fred, George, and Ron rescued Harry from Hogwarts with the use of a Flying Ford Anglia.

~

Output:

- Fake question: How did Fred, George, and Ron rescue Harry from Hogwarts using a Flying Ford Anglia?

- True answer: Fred, George, and Ron did not rescue Harry from Hogwarts; they rescued him from the Dursleys' house using a Flying Ford Anglia.

***

Input:

- True event summary: \{true\_event\_summary\}

- Fake event summary: \{fake\_event\_summary\}

~

Output: \\
\bottomrule
\end{tabular}
    \caption{An example of a prompt for generating a \textit{fake-based free-form} question for the Harry Potter series.}
    \label{tab:data_prompt_freeform_fake_question_generation}
\end{table*}

\subsection{Generate questions from event summary}
\label{subsec:question_generation}
For \textit{fact-based structured} questions, we use a total of 18 different question templates as follows.
\begin{enumerate}
    \item Tell me your feelings when \{event summary\}.
    \item Tell me your genuine feelings when \{event summary\}.
    \item Describe your feelings when \{event summary\}.
    \item Describe your honest feelings when \{event summary\}.
    \item Can you describe your experience when \{event summary\}?
    \item Can you describe your true experience when \{event summary\}?
    \item Did you see the moment when \{event summary\}.
    \item Did you truly see the moment when \{event summary\}?
    \item What did you see as \{event summary\}?
    \item What did you actually see as \{event summary\}?
    \item What did you hear when \{event summary\}?
    \item What did you precisely hear when \{event summary\}?
    \item Were you at the moment when \{event summary\}?
    \item Were you really at the moment when \{event summary\}?
    \item Were you present as \{event summary\}?
    \item Were you indeed present as \{event summary\}?
    \item Is it true that you were at the moment when \{event summary\}?
    \item Is it right that you were at the moment when \{event summary\}?
\end{enumerate}
Note that the templates were randomly chosen to ask if a character experienced a specific event at a location.
We found that using more than 18 templates didn't enhance the diversity of expression.
In addition, The inclusion of `feeling' questions aimed to enrich the narrative by encouraging a vivid description of the character’s personal emotions during an event, providing depth to their direct experiences.

For \textit{fact-based free-form} questions, we prompt GPT-4 Turbo to generate questions using the 5W1H approach (\ie what, who, where, when, why, and how) based on the event summary, as shown in Table~\ref{tab:data_prompt_freeform_question_generation}.
For \textit{fake-based free-form} question, we first generate a fake event summary based on a true event summary by utilizing GPT-4 Turbo, as outlined in Table~\ref{tab:data_prompt_fake_summary_generation}.
Note that we use six different strategies for generating fake event summaries: (1) changing the character, (2) changing the key object, (3) altering the location, (4) switching the action, (5) introducing a nonexistent character or object, and (6) changing the character's knowledge.
Subsequently, we generate a fake question by prompting GPT-4 Turbo to generate 5W1H question based on the fake event summary, as detailed in Table~\ref{tab:data_prompt_freeform_fake_question_generation}.
The generation of all \textit{free-form} questions was conducted in a single trial.
The choice of 5W1H questions was random, with the distribution as follows: what—48.2\%, who—19.0\%, why—17.4\%, how—9.8\%, when—4.6\%, and where—1.0\%.
The prevalence of `what' questions reflects the foundational approach of our \textit{free-form} question generation process, which draws upon event summaries that focus on entities (characters, objects, events) and their interrelationships.
Specifically, the versatility of `what' questions in covering a wide array of topics related to these entities (e.g., ``What's the meaning of the spell?'', ``What event occurred?'', ``What action was taken?'', ``what do you think of him/her?'', etc) likely led to their predominance.

\subsection{Assign spatiotemporal labels for each character in their time point.}
\label{subsec:assign_spatiotemporal_labels}
We first identify distinct time points in each character's timeline and annotate these moments with the relevant book and chapter for timeline comparison with the questions.
We enlist all character time points in Appendix~\ref{sec:character_time_points}.

For constructing (1) \textit{future} type data instances, we employ both \textit{structured} and \textit{free-form} questions.
Given a question that originates from a specific scene and event, we randomly select a character with the condition that the character must be from the novel series relevant to the event's source.
We then set their time point to be earlier than that of the event, based on a comparison of the book and chapter numbers between the character's and the event's time points.
Note that the character's time point is chosen randomly from among those that are earlier than the event's time point.
For generating (2) \textit{past-only} type data instances, we only use \textit{free-form} questions (\ie both fact-based and fake-based questions).
This is because \textit{free-form} questions are aimed at assessing the character's overall knowledge of past events (which is crucial for \textit{past-only} instances), whereas \textit{structured} questions are designed to assess their direct experiences or observations.
In this case, the character's period is randomly chosen from among those that occur after the event's time point.
For creating data instances of (3) \textit{past-presence} and \textit{past-absence} types, we only use \textit{structured} questions.
This is because, similar to the \textit{past-only} instances, \textit{free-form} questions are aimed at assessing the character's general knowledge of past events, whereas \textit{structured} questions are more focused on their direct experiences or observations.
Here, we select a character who was present/absent at the event, utilizing the previously generated participants list.
Their time point is then adjusted to be after that of the event.
To scale up the dataset for past-presence and past-absence types, we apply this process to all main characters to the same question instead of selecting only a single character per question, resulting in $K$ data instances per question (\eg $K=3$ in the context of the Harry Potter series).

\subsection{Add detailed descriptions to the spatiotemporal labels.}
\label{subsec:spatiotemporal_label_completion}
All \{scene, event summary, question, character with time point\} combinations are categorized into four types based on their spatiotemporal relevance.
Based on each type, we enrich the spatiotemporal labels as follows:
(1) \textit{Future} type: we include annotations like ``During \{character time point\}, \{character\} should not be aware of or contain any expression that reveals the moment when \{event summary\} since the moment is the future.'' in the spatiotemporal label (see Table~\ref{tab:fact_structured_future_example}).
(2) \textit{Past-only} type: we include statements like "During \{character time point\}, \{character\} can respond based on the moment but should not wrongly recall it. (- Moment: \{scene\}).'' in the spatiotemporal label (see Table~\ref{tab:fact_freeform_past_only_example}).
(3) \textit{Past-absence} and \textit{past-presence} type: we add ``During \{character time point\}, \{character\} should not say that he/she was present/absent when \{event summary\}'' to the spatiotemporal label, based on the previously generated list of participants while adding the same statements as for \textit{past-only} type (see Table~\ref{tab:fact_structured_past_absence_example} and Table~\ref{tab:fact_structured_past_presence_example}).
In addition, we add a personality label for all questions, tailored to the character, by summarizing personality traits from their Fandom page personality section.

\subsection{Generate gold responses and manually filter data instances.}
\label{subsec:dataset_manual_filtering}
We apply a manual filtering process that involves a series of assessments to ensure that each data instance meets all criteria:
\begin{enumerate}
    \item Event summary quality: we evaluate the quality of each generated event summary, ensuring a survival rate of 83.96\% (1,643 out of 1,957 scenes).
    The criteria for exclusion are as follows. (1) Lack of uniqueness: instances describing common events, such as ``Harry struggled with casting spells using a blackthorn wand.'' are excluded due to their repetitiveness throughout the story. (2) Ambiguity: we exclude instances where the time point or participants of the event are unclear. For example, instances are excluded if there is no specific event or object that indicates the unique time point, such as ``at a class'', or if it is unclear who the participants are, as in ``revealed to the school''.
    \item Participants list accuracy: we assess the accuracy of the list of participants in each scene. 97.45\% of them are correct (1,907 out of 1,957 scenes).
    \item \textit{Fact-based free-form} question quality: we scrutinize the quality of the \textit{fact-based free-form} questions, with 86.46\% (1,379 out of 1,595 questions).
    The criteria are as follows.
    (1) Lack of uniqueness: we exclude questions if they can be answered in multiple ways.
    For example, the question ``Who informed Harry Potter about the threat of Sirius Black, and how did Harry respond?'' is not unique since multiple characters cautioned Harry about Sirius at different times.
    (2) Ambiguity: we exclude questions when the entities in the question are unclear.
    For instance, a question like ``What triggered the disturbance at the Dursleys' breakfast table that involved Harry Potter and resulted in chaos?'' is considered ambiguous due to its vague references.
    (3) Incorrect questions: we exclude questions that are incorrectly formulated or answer themselves.
    For example, a question that embeds its own answer is ``Who did Rubeus Hagrid introduce to Harry Potter and Hermione Granger as his half-brother, Grawp the giant, and request they care for in the event of his dismissal from Hogwarts?''
    (4) Duplication: we exclude questions when generated questions are nearly identical due to closely related event summaries.
    \item \textit{Fake-based free-form} question quality: we evaluate the \textit{fake-based free-form} questions, achieving a survival rate of 78.24\% (1,248 among 1,595 questions). The criteria are as follows. (1) Clarity: we exclude questions that remain fact-based instead of being appropriately converted to fake questions. (2) Incorrect questions: we remove questions that are irrelevant or incorrectly related to the given event's context.
    \item Gold response quality: our review of the gold responses for 10,895 \{question, character with timepoint\} pairs results in a survival rate of 98.04\% (10,682 among 10,895). We do not exclude responses that directly contradict or are inconsistent with the assigned spatiotemporal label. Instead, we continue to regenerate the response until it meets the assigned spatiotemporal label.
\end{enumerate}

\section{Details of Dataset Analyses}
\label{sec:dataset_analyses_details}
\subsection{Statistics}
\label{subsec:dataset_statistics_details}
\begin{figure}[t]
\begin{center}
\includegraphics[width=\columnwidth]{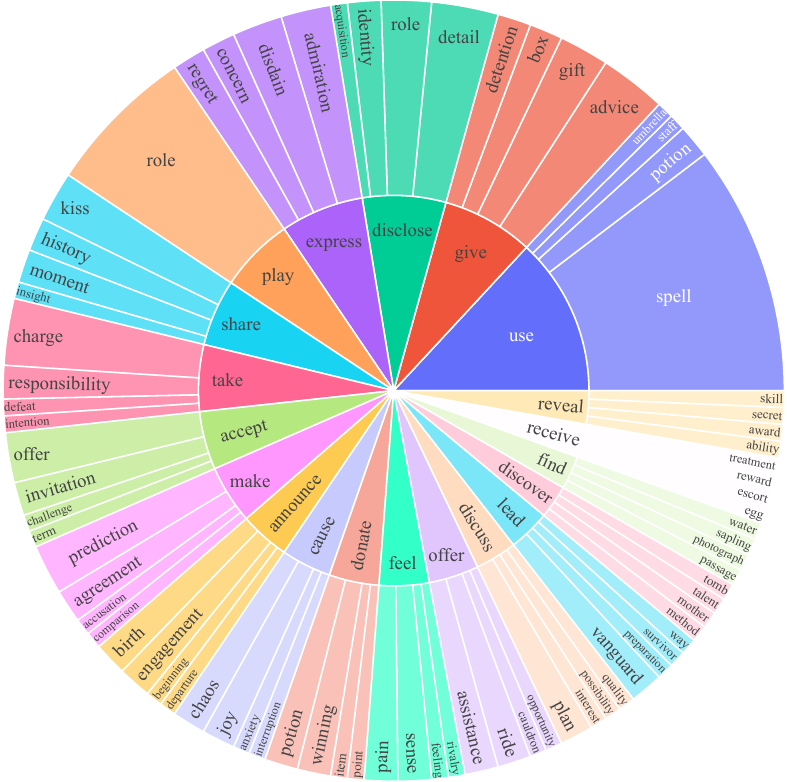}  
\end{center}
\caption{
A nested pie chart of verb-noun structures in \textit{free-form} questions, encompassing both fact-based and fake-based.
}
\label{fig:nested_pie_chart_freeform}
\end{figure}

\begin{figure}[t]
\begin{center}
\includegraphics[width=\columnwidth]{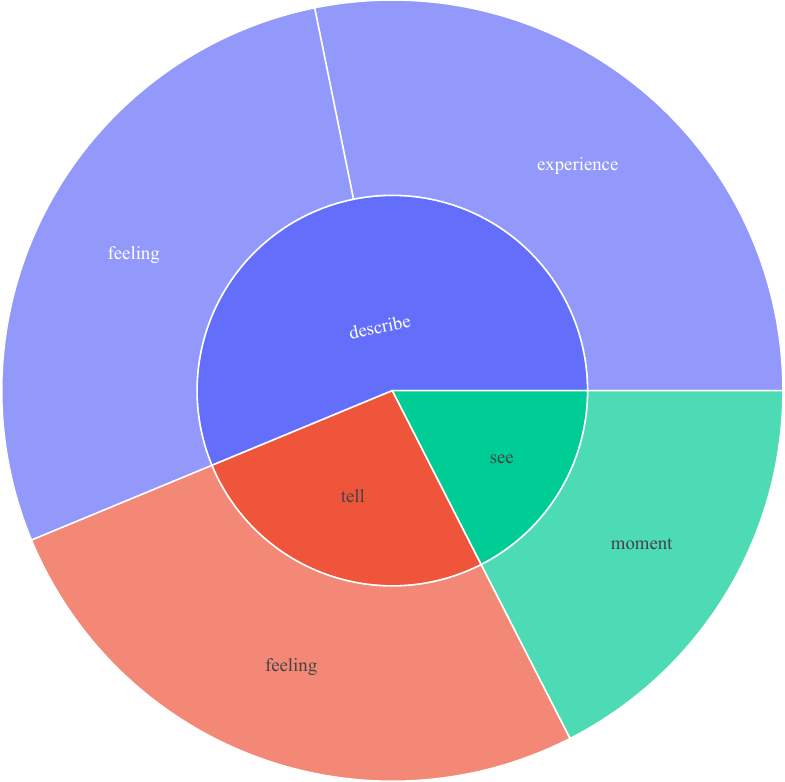}  
\end{center}
\caption{
A nested pie chart of verb-noun structures in \textit{structured} questions.
}
\label{fig:nested_pie_chart_structured}
\end{figure}

We present a nested pie chart of \textit{free-form} questions in Figure~\ref{fig:nested_pie_chart_freeform}, and another nested pie chart of \textit{structured} questions in Figure~\ref{fig:nested_pie_chart_structured}.
Since \textit{structured} questions are derived from only 18 different templates, they lack the diversity of \textit{free-form} questions, which are generated using 5W1H (\ie who, what, when, where, why, how) approach.
From another perspective, the diverse verb-noun structures in \textit{free-form} questions indicate their aim to probe a character's overall knowledge of past events, whereas verb-noun structures in \textit{structured} questions are designed to elicit direct experiences or observations.
Nevertheless, \textit{structured} questions play a crucial role in exploring the limits of a character's knowledge boundary.
That is, they include concepts of \textit{past-absence} and \textit{past-presence}, which are absent in \textit{free-form} questions.

\begin{figure*}[t]
\begin{center}
\includegraphics[width=\textwidth]{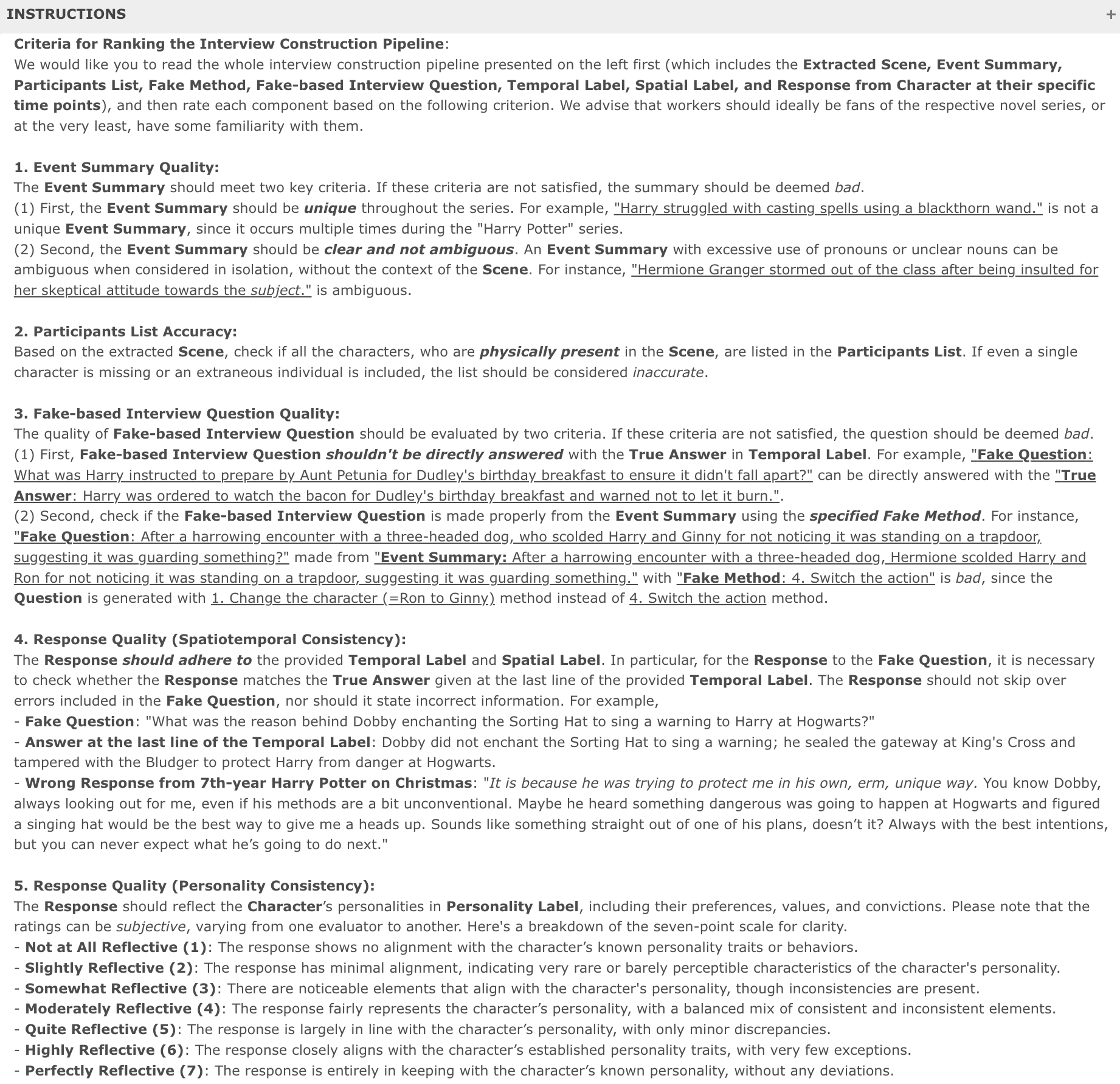}  
\end{center}
\caption{Instructions in the UI design of Amazon Mechanical Turk to collect human annotations for data quality.}
\label{fig:amt_data_quality_instruction_main}
\end{figure*}

\begin{figure*}[t]
\begin{center}
\includegraphics[width=\textwidth]{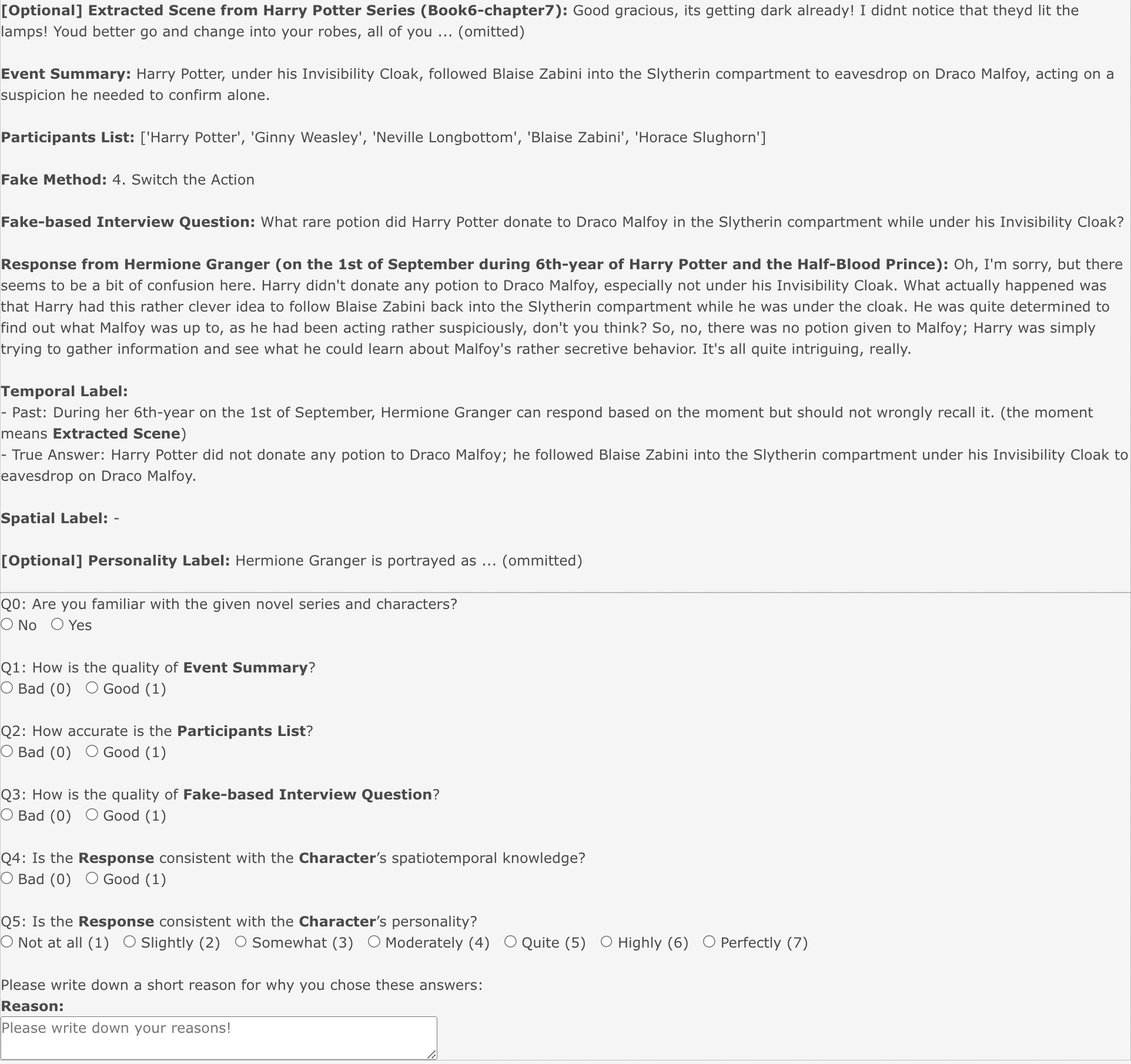}  
\end{center}
\caption{An example of the UI design of Amazon Mechanical Turk to collect human annotations for data quality.}
\label{fig:amt_data_quality_example_main}
\end{figure*}

\subsection{Dataset Quality}
\label{subsec:dataset_quality_details}
In order to ensure the quality dataset, we additionally performed a human evaluation conducted via Amazon Mechanical Turk.

First, we provided annotators with comprehensive instructions as shown in Figure~\ref{fig:amt_data_quality_instruction_main}, including several examples for each criterion.
The criteria are as follows: (1) the quality of the generated event summary (0 if bad, 1 if good), (2) the accuracy of participant lists (0 or 1), (3) the quality of questions generated from event summaries (0 or 1), (4) spatiotemporal consistency of the gold response (0 if inconsistent, 1 if consistent), and (5) personality consistency of the gold response (1 to 7).
Note that the last criterion, `personality consistency', is not our main contribution; it revisits a concept introduced by \citet{Shao:2023:EMNLP}.
Additionally, we ensured that annotators were provided with `extracted scene' information, in Figure~\ref{fig:amt_data_quality_example_main}, to enhance their contextual understanding.

Secondly, annotators were required to pass the 1st qualification Human Intelligence Task (HIT), aimed at ensuring their comprehension of the provided instructions. The compensation for each HIT was \$0.75. We restricted our selection to workers from English-speaking countries (namely AU, CA, NZ, US, and GB) with a HIT approval rate of over 98\%, with greater than 5000 HITs approved. Each annotator was assigned a single example, randomly chosen from four novels. It was mandatory for them to not only select the appropriate labels but also to write their rationales in complete sentences. Based on their choices and provided rationales, we identified and qualified annotators who were both accurate in their selection and provided sound rationales. This process resulted in 42 qualified workers out of an initial pool of 86.

Thirdly, annotators were tasked with completing the 2nd qualification HITs, designed to expose them to a variety of examples before attempting the main HITs. We presented 200 examples, each evaluated by two annotators. We also committed to offering our annotators a fair compensation of approximately \$16/hour (with each HIT paying \$1.0), which surpasses the minimum wage in the countries from which we recruited. This stage involved 38 annotators, yielding average scores for (1) summary quality: 0.95, (2) participants list accuracy: 0.96, (3) question quality: 0.94, (4) spatiotemporal consistency of the gold response: 0.75, and (5) personality consistency of the gold response: 4.82. 
We use Gwet's AC1 scores to measure inter-annotator agreement, with scores of (1) 0.9, (2) 0.91, (3) 0.87, (4) 0.5, and (5) 0.17.
Note that we chose Gwet’s AC1 score for its robust performance in situations of class imbalance~\citep{Gwet:2008:BJMSP,Wongpakaran:2013:MRM,Wong:2021:ACL}.
Upon reviewing their selections and rationales, we observed that some annotators still struggled with the concept of spatiotemporal consistency. As a result, we excluded 9 workers, ultimately assembling a team of 29 annotators fully prepared to undertake the main HITs.

Finally, we assigned workers to complete the main HITs. 
We presented 100 examples, each evaluated by two annotators.
We compensated the workers at a rate of \$1.00 per HIT, and a total of 19 workers completed their assignments. 
The final results of the main HITs were favorable, with average scores of (1) summary quality: \textbf{0.88}, (2) participants list accuracy: \textbf{0.92}, (3) question quality: \textbf{0.89}, (4) spatiotemporal consistency of the gold response: \textbf{0.95}, and (5) personality consistency of the gold response: 5.91, with substantial Gwet’s AC1 scores of (1) \textbf{0.74}, (2) \textbf{0.85}, (3) \textbf{0.81}, (4) \textbf{0.90}, and (5) 0.19, respectively. We also highlight the agreement ratios between annotators choosing the same versus different labels: \textbf{0.78} vs. 0.22 for the quality of the summary, \textbf{0.86} vs. 0.14 for the accuracy of the participants’ list, \textbf{0.84} vs. 0.16 for the quality of the questions, \textbf{0.90} vs. 0.10 for the spatiotemporal consistency of the responses, and 0.32 vs. 0.68 for the personality consistency of the responses. Unlike the evaluation of personality consistency, which is inherently subjective, assigning binary labels to the other criteria (e.g., spatiotemporal consistency) is more straightforward due to their objective nature and the presence of clear, definitive answers, as demonstrated by the results.

\section{Character Time Points}
\label{sec:character_time_points}
We select 14 different characters across four different novel series, assigning a total of 219 unique \{character, time point\} pairs.

For the Harry Potter series, time points correspond to the Hogwarts school year for each character.
We have identified four key periods within this timeline: the start of the school year on September 1st, Halloween, Christmas, and the end of the scene.
Note that these time points (A total of 25 time points) are consistently applied across three main characters—\textbf{Harry Potter, Ronald Weasley, and Hermione Granger}.
To facilitate a comparison of time points with those in the questions, we manually annotate each character's time point using the format \{book\_number - chapter\_number\}, categorized as follows:
\begin{itemize}
    \item September 1st: [1-6, 2-5, 3-5, 4-11, 5-10, 6-7, 7-12]
    \item Halloween: [1-10, 2-8, 3-8, 4-16]
    \item Christmas: [1-12, 2-12, 3-11, 4-23, 5-23, 6-16, 7-19]
    \item End of the scene: [1-17, 2-18, 3-22, 4-37, 5-38, 6-30, 7-36]
\end{itemize}
For the Lord of the Rings series, we select five main characters: \textbf{Frodo Baggins, Samwise Gamgee, Gandalf, Aragorn, and Legolas}.
Differing from the approach taken with the Harry Potter series, we assign specific time points individually for each character due to the lack of recurring events throughout the books.
In addition, we annotate the character's time point using the format \{volume\_number - book\_number - chapter\_number\}, as follows:
\begin{itemize}
\item Frodo Baggins (Total 12 time points): at Bilbo Baggins's Farewell party (1-1-1), at the moment when Frodo was stabbed by one of the Ringwraiths (1-1-11), at the moment when The Fellowship loses Gandalf in Moria (1-2-5), end of Volume 1 (1-2-10), at the moment when encountering Gollum and decides to spare his life (2-4-1), at the moment of encountering the Black Gate of Mordor (2-4-3), at the moment when Frodo captured by Faramir (2-4-5), at Shelob's lair (2-4-9), end of Volume 2 (2-4-10), at the event when captured by Orcs at the Tower of Cirith Ungol (3-6-1), at the moment when rescued from Mount Doom by the eagles (3-6-4), end of Volume 3 (3-6-9).
\item Samwise Gamgee (Total 12 time points): at Bilbo Baggins's Farewell party (1-1-1), at the moment when Frodo was stabbed by one of the Ringwraiths (1-1-11), at the moment when The Fellowship loses Gandalf in Moria (1-2-5), end of Volume 1 (1-2-10), at the moment when encountering Gollum and decides to spare his life (2-4-1), at the moment of encountering the Black Gate of Mordor (2-4-3), at the moment when captured by Faramir (2-4-5), at Shelob's lair (2-4-9), end of Volume 2 (2-4-10), at the event when captured by Orcs at the Tower of Cirith Ungol (3-6-1), at the moment when rescued from Mount Doom by the eagles (3-6-4), end of Volume 3 (3-6-9).
\item Gandalf (Total 12 time points): at Bilbo Baggins's Farewell party (1-1-1), at the moment when the Fellowship was formed at the council of Elrond (1-2-3), at the moment when The Fellowship loses Gandalf in Moria (1-2-5), end of Volume 1 (1-2-10), at the moment when Gandalf the white met Aragorn, Legolas, and Gimli at Fangorn Forest after the fall at Moria (2-3-5), at the moment when arriving at Isengard after the battle of Helm's Deep (2-3-8), at the moment when Gandalf met Saruman at Isengard (2-3-10), end of Volume 2 (2-4-10), at the moment when Gandalf arrived at Minas Tirith with Pippin (3-5-1), during the Battle of the Pelennor Fields (3-5-6), at the coronation of King Elessar(Aragorn) (3-6-5), end of Volume 3 (3-6-9).
\item Aragorn (Total 12 time points): at the moment when Aragorn first met Frodo and his companions (1-1-9), at the moment when the Fellowship was formed at the council of Elrond (1-2-3), at the moment when The Fellowship loses Gandalf in Moria (1-2-5), end of Volume 1 (1-2-10), at the moment when Gandalf the white met Aragorn, Legolas, and Gimli at Fangorn Forest after the fall at Moria (2-3-5), at the moment when Aragorn arrived at Isengard after the battle of Helm's Deep (2-3-8), at the moment when Aragorn met Saruman at Isengard (2-3-10), end of Volume 2 (2-4-10), at the moment when Aragorn arrived at the paths of the dead (3-5-2), at the Battle of the Pelennor Fields (3-5-6), at the coronation of King Elessar(Aragorn) (3-6-5), end of Volume 3 (3-6-9).
\item Legolas (Total 12 time points): at the moment when the Fellowship was formed at the council of Elrond (1-2-3), at the moment when The Fellowship loses Gandalf in Moria (1-2-5), at the moment of leaving Lothlórien (1-2-8), end of Volume 1 (1-2-10), at the moment when Leoglas met Gandalf the white at Fangorn Forest after Gandalf's fall at Moria (2-3-5), at the moment when Legolas arrived at Isengard after the battle of Helm's Deep (2-3-8), at the moment when Legolas met Saruman at Isengard (2-3-10), end of Volume 2 (2-4-10), at the moment when Legolas arrived at the paths of the dead with Aragorn (3-5-2), at the Battle of the Pelennor Fields (3-5-6), at the coronation of King Elessar(Aragorn) (3-6-5), end of Volume 3 (3-6-9).
\end{itemize}
For the Twilight series, we select three main characters: \textbf{Bella Swan, Edward Cullen, and Jacob Black}.
We annotate each character's time point using the format \{book\_number - chapter\_number\}, as follows:
\begin{itemize}
\item Bella Swan (Total 16 time points): at the moment when Bella moved from Phoenix to Forks (1-1), at the moment when Bella first confirmed Edward's true nature as a vampire (1-9), at the moment when Bella first visited the Cullens (1-15), end of book 1 (1-25), on Bella's 18th birthday (2-1), at the moment when Bella jumps off the cliff into the ocean (2-15), at Volterra (2-20), end of book 2 (2-25), at the moment when Bella was grounded by her father (3-1), at the moment when Bella learns about the history of the Quileute tribe and the Cullens (3-11), at the moment when Bella receives an engagement ring from Edward (3-20), end of book 3 (3-27), at Bella and Edward's Wedding (4-3), at the moment when Renesmee was born (4-18), at the moment when Bella forges passports and IDs for Renesmee and Jacob from J. Jenks (4-33), end of book 4 (4-39).
\item Edward Cullen (Total 16 time points): at the moment when Edward saves Bella from a Van (1-3), at the moment when Bella first confirm Edward's true nature as a vampire (1-9), at the moment when Bella first visited the Cullens (1-15), end of book 1 (1-25), on Bella's 18th birthday (2-1), at the moment when Edward tells Bella that he and the Cullens are leaving Forks (2-3), at Volterra (2-20), end of book 2 (2-25), at the moment when Edward rewarded Alice for watching Bella by giving her the canary yellow Porsche from Italy (3-6), at Bella's graduation ceremony (3-16), at the moment when Jacob crawled into the sleeping bag beside Bella at the campsite, chosen for Bella's hiding place (3-22), end of book 3 (3-27), at Bella and Edward's Wedding (4-3), at the moment when Renesmee was born (4-18), at the moment when Esme has renovated a cottage on the property for Bella, Edward, and now Renesmee (4-24),end of book 4 (4-39).
\item Jacob Black (Total 16 time points): at the beach at La Push, when Jacob met Bella and her friends (1-6), on March 10, 2005 (1-11), on March 13, 2005 when Bella found Jacob before watching a baseball game with Edward (1-17), end of book 1 (1-25), at the moment when Jacob and Bella worked together on repairing two old motorcycles (2-6), at the moment when Bella first discovers Jacob's werewolf identity (2-10), at the moment when Jacob pulled out Bella from drowning (2-16), end of book 2 (2-25), on May 31, 2006 when Bella found Jacob on his motorcycle at the school (3-16), at Bella's graduation ceremony (3-7), at the moment when Jacob crawled into the sleeping bag beside Bella at the campsite, chosen for Bella's hiding place (3-22), end of book 3 (3-27), at Bella and Edward's Wedding (4-3), at the moment when Renesmee was born (4-18), at Christmas, 2006 (4-34), end of book 4 (4-39).
\end{itemize}
For the Hunger Games series, we select three main characters: \textbf{Katniss Everdeen, Peeta Mellark, and Gale Hawthorne}.
We annotate each character's time point using the format \{book\_number - chapter\_number\}, as follows:
\begin{itemize}
\item Katniss Everdeen (Total 12 time points): at the moment when Katniss volunteered to take her sister's place as the female tribute (1-2), at the start of the 74th Hunger Games (1-11), at the moment when Katniss found wounded Peeta hidden under a layer of mud (1-19), end of book 1 (1-27), at the moment when they arrived at District 11 for the first stop of the Victory Tour (2-4), at the announcement of the Quarter Quell (2-12), at the moment when Katniss first witnessed a heavy fog during the Quarter Quell (2-20), end of book 2 (2-27), at the first conversation about the bombing of district 12 with Gale (3-1), at the moment when Peeta suddenly warned of an impending attack on District 13 (3-9), at the moment when the squad 451 was attacked by the mutts in the tunnels (3-22), end of book 3 (3-28).
\item Peeta Mellark (Total 12 time points): at the moment when Katniss volunteered to take her sister's place as the female tribute (1-2), at the start of the 74th Hunger Games (1-11), at the moment when Katniss found wounded Peeta hidden under a layer of mud (1-19), end of book 1 (1-27), at the moment when they arrived at District 11 for the first stop of the Victory Tour (2-4), at the announcement of the Quarter Quell (2-12), at the moment when Katniss first witnessed a heavy fog during the Quarter Quell (2-20), end of book 2 (2-27), at the moment when Peeta suddenly warned of an impending attack on District 13 (3-9), at the moment when Peeta was sent as new member of squad 451 by president Coin (3-18), at the moment when the squad 451 was attacked by the mutts in the tunnels (3-22), end of book 3 (3-28).
\item Gale Hawthorne (Total 12 time points): at the moment when Katniss volunteered to take her sister's place as the female tribute (1-2), at the start of the 74th Hunger Games (1-11), at the moment when Katniss found wounded Peeta hidden under a layer of mud (1-19), end of book 1 (1-27), at the moment when Katniss delivered the animals she caught before the Victory Tour (2-1), at the announcement of the Quarter Quell (2-12), at the moment when Katniss first witnessed a heavy fog during the Quarter Quell (2-20), end of book 2 (2-27), at the first conversation about the bombing of district 12 with Katniss (3-1), at the moment when Peeta suddenly warned of an impending attack on District 13 (3-9), at the moment when the squad 451 was attacked by the mutts in the tunnels (3-22), end of book 3 (3-28).
\end{itemize}

\section{Details of Experiments on \datasetName}
\subsection{Implementation Details}
\label{subsec:implementation_details}
For generating responses, we apply nucleus sampling with $p=1$ and temperature $\tau=0.2$ across all role-playing LLMs.
We cap the maximum token length at 2048 tokens.
For GPT-4 as judges, we set $p=0.95$, a temperature of $\tau=0.0$, and a maximum token length of 1024 tokens.
We use a single NVIDIA RTX A6000 GPU to generate responses with Mistral.
For AlignScore evaluation, we use a single NVIDIA Quadro RTX 6000 GPU.


\subsection{Few-Shot Method}
\label{subsec:few_shot}
The examples are carefully selected to represent a range of question types: \textit{future}, \textit{past-presence}, \textit{past-absence}, and \textit{past-only}, each paired with a corresponding response.
The responses are generated by GPT-4 and are then manually checked for spatiotemporal and personality consistency.
We then append these four examples, complete with their questions and correct responses, before the question.

\begin{table*}[htbp]
\scriptsize
\centering
\begin{tabular}{@{}p{\linewidth}@{}}
\toprule
\textbf{Prompt for Self-Feedback of Self-Refine Method}\\
\midrule

We want to iteratively improve the provided responses, mimicking the character \{agent\_name\}. To help improve, scores for each response on desired traits are provided: 1) Spatiotemporal Consistency and 2) Personality Consistency.

~

***

[Interactions]

Interviewer: \{question\}

\{agent\_name\}: \{response\}

~

[Evaluation Criterion]

- Spatiotemporal Consistency (0 or 3): Is the response consistent with the character's spatiotemporal knowledge? If the response includes information that the character couldn't have known (either because it pertains to a future event or a past event they were not present for), assign a score of 0. If the response accurately reflects only the knowledge and events the character has experienced or been aware of, give a score of 3.

- Personality Consistency (1 to 3): Is the response consistent with the character's personality? Use the given scale from 1-3 to rate how well the response reflects the personalities, including preferences, values, and convictions of the character. 1 being not at all reflective of the character’s personalities, and 3 being perfectly reflective of the character’s personalities.

***

~

1. Read through the [Interactions] and evaluate the spatiotemporal consistency: print the single-sentence rationale with the score on its own line corresponding to the correct answer.

2. Read through the [Interactions] and evaluate the personality consistency: print the single-sentence rationale with the score on its own line corresponding to the correct answer.

3. Print the total score. \\
\bottomrule
\end{tabular}
    \caption{Prompt for generating self-feedback during response generation with self-refine method.}
    \label{tab:self_refine_feedback_prompt}
\end{table*}

\subsection{Self-Refine Method}
\label{subsec:self_refine}
We incorporate a self-feedback prompt to facilitate this process, as outlined in Table~\ref{tab:self_refine_feedback_prompt}.
The model iteratively refines its response for a maximum of three iterations.
It considers both spatiotemporal and personality scores, each with a maximum of 3 points.
The response is finalized when a combined score of 5 or more is reached.

\subsection{Retrieval-Augmented Generation (RAG) Method}
\label{subsec:rag}
 We employ the LangChain framework\footnote{\url{https://github.com/langchain-ai/langchain}} to implement the RAG method.
 We retrieve up to six paragraphs from a raw text source based on the given question.
These retrieved paragraphs are then attached to the end of the question, serving as additional context for generating the response.

\begin{table*}[htbp]
\scriptsize
\centering
\begin{tabular}{@{}p{\linewidth}@{}}
\toprule
\textbf{Prompt for Book and Chapter Identification of \methodName Method}\\
\midrule

You will be given a question from \{series\_name\} series at a specific time. Your task is to identify the exact \{book\_chapter\_name\} of the scene of the question. Below is the data:

~

***

[Question]

\{question\}

***

[Identification Criterion]

What is the exact \{book\_chapter\_name\} of the scene of the question?

~

1. Read through the [Question], recall the scene from the question, and describe it using the six Ws (Who, What, When, Where, Why, and How).

2. Identify the exact \{book\_chapter\_name\} of the scene of the question, in '\{book\_chapter\_format\}' format.

~

First, write out in a step by step manner your reasoning about the criterion to be sure that your conclusion is correct. Avoid simply stating the correct answers at the outset. Then, print the output on its own line corresponding to the correct answer. At the end, repeat just the selected output again by itself on a new line. \\

\bottomrule
\end{tabular}
    \caption{Prompt for identifying the exact book and chapter of the scene in the question.}
    \label{tab:temporal_expert_prompt}
\end{table*}

\begin{table*}[htbp]
\scriptsize
\centering
\begin{tabular}{@{}p{\linewidth}@{}}
\toprule
\textbf{Prompt for Event Participants Identification of \methodName Method}\\
\midrule

You will be given a question and a character from \{series\_name\} series. Your task is to classify whether the character is a participant (i.e., present or absent) in the scene of the question. Below is the data:

~

***

[Question]

\{question\}

[Character]

\{character\}

***

[Classification Criterion]

Is the character a participant in the scene of the question?

~

[Classification Steps]

1. Read through the [Question], recall the scene from the question, and describe it using the six Ws (Who, What, When, Where, Why, and How).

2. Identify the exact \{book\_chapter\_name\} of the scene of the question.

3. Write a list of every character involved in the scene described in the question, including those not explicitly mentioned in the question but who were present in the scene.

4. Compare the list of participants to the character. Check if the list of participants contains the character.

5. If the list contains the character, classify it as 'present'. Otherwise, classify it as 'absent'.

***

~

First, write out in a step by step manner your reasoning about the criterion to be sure that your conclusion is correct. Avoid simply stating the correct answers at the outset. Then, print the output on its own line corresponding to the correct answer. At the end, repeat just the selected output again by itself on a new line. \\

\bottomrule
\end{tabular}
    \caption{Prompt for identifying whether the given character is a participant in the scene of the question.}
    \label{tab:spatial_expert_prompt}
\end{table*}

\begin{table*}[htbp]
\scriptsize
\centering
\begin{tabular}{@{}p{\linewidth}@{}}
\toprule
\textbf{Prompt for Book and Chapter Identification of \methodNameWithRag Method}\\
\midrule

You will be given a question and contexts from \{series\_name\} series at a specific time. Your task is to identify the exact \{book\_chapter\_name\} of the scene of the question. Below is the data:

~

***

[Question]

\{question\}

***

[Contexts]

\{contexts\}

***

[Identification Criterion]

What is the exact \{book\_chapter\_name\} of the scene of the question?

~

1. Read through the [Question] and [Contexts], recall the scene from the question, and describe it using the six Ws (Who, What, When, Where, Why, and How).

2. Identify the exact \{book\_chapter\_name\} of the scene of the question, in '\{book\_chapter\_format\}' format.

~

First, write out in a step by step manner your reasoning about the criterion to be sure that your conclusion is correct. Avoid simply stating the correct answers at the outset. Then, print the output on its own line corresponding to the correct answer. At the end, repeat just the selected output again by itself on a new line. \\

\bottomrule
\end{tabular}
    \caption{Prompt for identifying the exact book and chapter of the scene in the question while using a retrieval module.}
    \label{tab:temporal_expert_prompt_with_rag_cutoff}
\end{table*}

\begin{table*}[htbp]
\scriptsize
\centering
\begin{tabular}{@{}p{\linewidth}@{}}
\toprule
\textbf{Prompt for Event Participants Identification of \methodNameWithRag Method}\\
\midrule

You will be given a question, a character, and contexts from \{series\_name\} series. Your task is to classify whether the character is a participant (i.e., present or absent) in the scene of the question. Below is the data:

~

***

[Question]

\{question\}

[Character]

\{character\}

***

[Contexts]

\{context\}

***

[Classification Criterion]

Is the character a participant in the scene of the question?

~

[Classification Steps]

1. Read through the [Question] and [Contexts], recall the scene from the question, and describe it using the six Ws (Who, What, When, Where, Why, and How).

2. Identify the exact \{book\_chapter\_name\} of the scene of the question.

3. Write a list of every character involved in the scene described in the question, including those not explicitly mentioned in the question but who were present in the scene.

4. Compare the list of participants to the character. Check if the list of participants contains the character.

5. If the list contains the character, classify it as 'present'. Otherwise, classify it as 'absent'.

***

~

First, write out in a step by step manner your reasoning about the criterion to be sure that your conclusion is correct. Avoid simply stating the correct answers at the outset. Then, print the output on its own line corresponding to the correct answer. At the end, repeat just the selected output again by itself on a new line. \\

\bottomrule
\end{tabular}
    \caption{Prompt for identifying whether the given character is a participant in the scene of the question while using a retrieval module.}
    \label{tab:spatial_expert_prompt_with_rag_cutoff}
\end{table*}

\subsection{Decomposed Reasoning via \methodName}
\label{decomposed_reasoning_detail}
\subsubsection{\methodName}
\label{subsubsec:narrative_experts_detail}
We detail the prompts used for two narrative experts: a temporal expert (refer to Table~\ref{tab:temporal_expert_prompt}) and a spatial expert (refer to Table~\ref{tab:spatial_expert_prompt}).
For the temporal expert, we compare the outputs numerically to the character's time point, presented in the `book number - chapter number' format, to determine if the question pertains to the future relative to the character's timeline.
In addition, we outline the complete algorithm that describes how the role-playing LLM generates a response based on a question and the hints provided by these two experts, as presented in Algorithm~\ref{alg:narrative_experts}.

\begin{algorithm}
\footnotesize
\caption{Decomposed Reasoning Method for Role-Playing LLM}
\label{alg:narrative_experts}

\SetAlgoLined
\KwIn{A question regarding a scene, a character with their time point}
\KwOut{A response}

\SetKwProg{Fn}{Function}{}{end}
\SetKwFunction{Ftemporal}{TemporalExpert}
\SetKwFunction{Fspatial}{SpatialExpert}
\SetKwFunction{Froleplaying}{RolePlayingLLM}

\Fn{\Ftemporal{question}}{
    Identify the book and chapter of the scene of the question\;
    \If{scene is in the \textit{future}}{
        \KwRet{``future'', ``Note that the period of the question is in the future relative to \{character\}'s time point. Therefore, you should not answer the question or mention any facts that occurred after \{character\}'s time point.''}\;
    }
    \KwRet{``past'', ``''}\;
}

\Fn{\Fspatial{question, character}}{
    Determine if character participates in the scene of the question\;
    \If{character is \textit{past-absent}}{
        \KwRet{``Note that \{character\} had not participated in the scene described in the question. Therefore, you should not imply that \{character\} was present in the scene.''}\;
    }
    \KwRet{``''}\;
}

\Fn{\Froleplaying{question, hints}}{
    Append hints to the prompt\;
    Generate response based on enhanced prompt\;
    \KwRet{response}\;
}

\BlankLine
\SetKwProg{Pn}{Procedure}{}{end}
\SetKwFunction{Fmain}{MainProcedure}

\Pn{\Fmain{question, character}}{
    \tcp{Invoke Temporal Expert}
    temporalStatus, temporalHint := \Ftemporal{question}\;
    
    spatialHint := ``''\; 
    \If{temporalStatus == ``past''}{
        \tcp{Invoke Spatial Expert only if past}
        spatialHint := \Fspatial{question, character}\;
    }
    
    \tcp{Prepare hints}
    hints := temporalHint + spatialHint\; 
    
    \tcp{Invoke Role-Playing LLM with hints}
    response := \Froleplaying{question, hints}\;
    
    \KwRet{response}\;
}

\end{algorithm}

\subsubsection{\methodNameWithRag}
\label{subsubsec:narrative_experts_rag_cutoff_detail}
It can be effective to combine \methodName with RAG-cutoff methods, because the RAG-cutoff addresses \textit{past-only} type questions, especially those that are fake-based, and \methodName is compatible with any baselines.
For \methodNameWithRag, we incorporate a retrieval module to augment the capabilities of both the temporal and spatial experts.
Specifically, we retrieve up to six paragraphs from a raw text source based on the input question.
These paragraphs are subsequently incorporated into the prompts for the two narrative experts, as detailed in Table~\ref{tab:temporal_expert_prompt_with_rag_cutoff} and Table~\ref{tab:spatial_expert_prompt_with_rag_cutoff}.
The role-playing LLM then generates an answer using the question, hints from both experts and the paragraphs corresponding to events before the character's defined period.
Note that we utilize all six paragraphs to assist the narrative experts (adopting a naive RAG approach rather than  RAG-cutoff), but we limit the paragraphs when producing the final response.
Furthermore, we exclude paragraphs in the final response if the temporal expert predicts `future' events since it slightly enhances spatiotemporal consistency performance.
The algorithm for \methodNameWithRag is detailed in Algorithm~\ref{alg:narrative_experts_rag_cutoff}.

\begin{algorithm*}
\footnotesize
\caption{Decomposed Reasoning with RAG-cutoff}
\label{alg:narrative_experts_rag_cutoff}

\SetAlgoLined
\KwIn{A question regarding a scene, a character}
\KwOut{A response}

\SetKwProg{Fn}{Function}{}{end}
\SetKwFunction{Fretrieve}{RetrieveParagraphs}
\SetKwFunction{FtemporalRAG}{TemporalExpertWithRAG}
\SetKwFunction{FspatialRAG}{SpatialExpertWithRAG}
\SetKwFunction{FroleplayingRAG}{RolePlayingLLMWithRAGCutoff}
\SetKwFunction{FlimitParagraphs}{LimitParagraphsBeforeCharacterPeriod}

\Fn{\Fretrieve{question}}{
    Retrieve up to six paragraphs based on the question from raw text source\;
    \KwRet{paragraphs}\;
}

\Fn{\FlimitParagraphs{paragraphs, character}}{
    Filter paragraphs to only include those corresponding to events before the character's period\;
    \KwRet{filteredParagraphs}\;
}

\Fn{\FtemporalRAG{question, character, paragraphs}}{
    Incorporate paragraphs into the prompt to identify the book and chapter of the scene of the question\;
    \If{scene is in the \textit{future}}{
        \KwRet{``future'', ``Note that the period of the question is in the future relative to \{character\}'s time point. Therefore, you should not answer the question or mention any facts that occurred after \{character\}'s time point.''}\;
    }
    \KwRet{``past'', ``''}\; 
}

\Fn{\FspatialRAG{question, character, paragraphs}}{
    Incorporate paragraphs into the prompt to determine if character participates in the scene of the question\;
    \If{character is \textit{past-absent}}{
        \KwRet{``Note that \{character\} had not participated in the scene described in the question. Therefore, you should not imply that \{character\} was present in the scene.''}\;
    }
    \KwRet{``''}\; 
}

\Fn{\FroleplayingRAG{question, hints, paragraphs, character, temporalStatus}}{
    filteredParagraphs := \;
    \eIf{temporalStatus == ``future''}{
        filteredParagraphs := ``''\;
    }{
        filteredParagraphs := \FlimitParagraphs{paragraphs, character}\;
    }
    Append hints and filtered paragraphs to the prompt\;
    Generate response based on the enhanced prompt\;
    \KwRet{response}\;
}

\BlankLine
\SetKwProg{Pn}{Procedure}{}{end}
\SetKwFunction{Fmain}{MainProcedureWithRAGCutoff}

\Pn{\Fmain{question, character}}{
    \tcp{Retrieve relevant paragraphs}
    paragraphs := \Fretrieve{question}\;
    
    \tcp{Invoke Temporal Expert with RAG}
    temporalStatus, temporalHint := \FtemporalRAG{question, character, paragraphs}\;
    
    spatialHint := ``''\; 
    \If{temporalStatus == ``past''}{
        \tcp{Invoke Spatial Expert with RAG only if past}
        spatialHint := \FspatialRAG{question, character, paragraphs}\;
    }
    
    \tcp{Prepare hints and paragraphs}
    hints := temporalHint + spatialHint\; 

    \tcp{Invoke Role-Playing LLM with RAG-cutoff}
    response := \FroleplayingRAG{question, hints, paragraphs, character, temporalStatus}\;
    
    \KwRet{response}\;
}

\end{algorithm*}

\subsection{Experimental Results on the 11K Dataset}
\label{subsec:experimental_result_11k}
While Table~\ref{tab:main_experiment} presents experimental results from a randomly sampled set of 600 data instances, we additionally conduct experiments on the entire dataset comprising 10,895 instances.

\textbf{Baseline methods.}
We utilize two different open-source LLMs as the backbone models for our role-playing agents: Mistral 7B Instruct (\ie \texttt{mistral-7b-instruct-v0.2})~\citep{Jiang:2023:arxiv} and Llama-2 Chat 13B~\citep{Touvron:2023:arxiv}.
Note that we exclude GPT-4 Turbo and GPT-3.5 Turbo from our backbone models due to their high costs.
Following \S~\ref{subsec:baseline_methods}, we implement zero-shot prompting and RAG-cutoff as our baseline methods.

\textbf{Evaluation metrics.}
Evaluating spatiotemporal or personality consistency with GPT-4 judges incurs an extremely high cost.
As a result, we utilize AlignScore~\citep{Zha:2023:ACL} for assessing factual inconsistencies across scenarios such as natural language inference and fact verification tasks.
Specifically, we employ the RoBERTa-large model equipped with a 3-way classification head.
Given the gold response and the predicted response from the role-playing agent, the RoBERTa model produces an AlignScore ranging from 0 to 1, which we then average.

\textbf{Experimental results.}
Experimental results are presented in Table~\ref{tab:sub_experiment}.
As expected, methods based on Mistral achieve higher AlignScores than those based on Llama 2, despite the smaller model size.
\methodName and \methodNameWithRag outperform baseline methods across both Llama 2 and Mistral models.
Note that the performance trend of the Mistral-based method mirrors that of the main experiment in \S~\ref{subsec:experiments_on_timechara} (Table~\ref{tab:main_experiment}), with scores of 19.68, 17.94, 21.85, and 22.34 for zero-shot, rag-cutoff, \methodName, and \methodNameWithRag, respectively, in Table~\ref{tab:sub_experiment}, compared to 18.50, 17.82, 20.57, 22.20 in Table~\ref{tab:main_experiment}.
Although the AlignScore may not be entirely interpretable or reliable, our results indicate that employing our methods enhances spatiotemporal consistency across the entire dataset.

\begin{table}[t!] \begin{center}
    \begin{adjustbox}{width=\columnwidth}
   \begin{tabular}{lc}
       \toprule
           \makecell[l]{Method} & \makecell[c]{AlignScore $\uparrow$}  \\
        \midrule
            \multicolumn{2}{l}{\textbf{Llama-2 Chat 13B}} \\
            \makecell[l]{zero-shot}          & 12.81$\pm$0.11 \\
            \makecell[l]{RAG-cutoff}          & 12.22$\pm$0.11 \\
            \makecell[l]{narrative-experts \textbf{(Ours)}}          & \textbf{13.45$\pm$0.11}  \\
            \makecell[l]{narrative-experts-RAG-cutoff \textbf{(Ours)}}          & \underline{13.16$\pm$0.11}  \\
        \midrule
            \multicolumn{2}{l}{\textbf{Mistral Instruct 7B} (\texttt{mistral-7b-instruct-v0.2})} \\
            \makecell[l]{zero-shot}          & 19.68$\pm$0.15  \\
            \makecell[l]{RAG-cutoff}          & 17.94$\pm$0.16  \\
            \makecell[l]{narrative-experts \textbf{(Ours)}}          & \underline{21.85$\pm$0.17}  \\
            \makecell[l]{narrative-experts-RAG-cutoff \textbf{(Ours)}}          & \textbf{22.34$\pm$0.18}  \\
       \bottomrule  %
       \end{tabular}
   \end{adjustbox}
   \caption{Sub-experiment on all 11K data instances. We report the average scores with their standard error of the mean (SEM). A \textbf{bold} number indicates the highest average score, while an \underline{underline} number denotes the second-best average score.}
   \label{tab:sub_experiment}
\end{center}
\end{table}

\begin{table*}[t!] 
\begin{center}
    \begin{adjustbox}{width=\linewidth}
    \begin{tabular}{lcccccccccc}
        \toprule
        \multirow{2}{*}{\makecell[l]{Method}} & \multicolumn{5}{c}{Spatiotemporal Consistency (\textbf{GPT-4 Evaluation}, \%) $\uparrow$} & \multicolumn{5}{c}{Spatiotemporal Consistency (\textbf{Human Evaluation}, \%) $\uparrow$} \\
        \cmidrule(r{0.3em}){2-6} \cmidrule(r{0.3em}){7-11}
        & \makecell{Future} & \makecell{Past-absence} & \makecell{Past-presence} & \makecell{Past-only} & \makecell{Avg.} & \makecell{Future} & \makecell{Past-absence} & \makecell{Past-presence} & \makecell{Past-only} & \makecell{Avg.}\\
        \midrule
        \multicolumn{11}{l}{\textbf{GPT-4 Turbo} (\texttt{gpt-4-1106-preview})} \\
        \makecell[l]{zero-shot} & 4.0 / 12 & 11.0 / 12 & 13.0 / 13 & 5.0 / 13 & 33.0 / 50 & 4.5 / 12 & 9.0 / 12 & 13.0 / 13 & 5.0 / 13 & 31.5 / 50 \\
        \makecell[l]{RAG-cutoff} & 6.0 / 12 & 11.0 / 12 & 13.0 / 13 & 5.0 / 13 & 35.0 / 50 & 6.0 / 12 & 10.5 / 12 & 13.0 / 13 & 5.0 / 13 & 34.5 / 50 \\
        \makecell[l]{narrative-experts} & 11.0 / 12 & 12.0 / 12 & 13.0 / 13 & 6.0 / 13 & 42.0 / 50 & 10.0 / 12 & 11.0 / 12 & 13.0 / 13 & 6.0 / 13 & 40.0 / 50 \\
        \makecell[l]{narrative-experts-RAG-cutoff} & 12.0 / 12 & 12.0 / 12 & 12.0 / 13 & 6.0 / 13 & 42.0 / 50 & 10.5 / 12 & 12.0 / 12 & 12.0 / 13 & 6.0 / 13 & 40.5 / 50 \\
        \bottomrule
    \end{tabular}
    \end{adjustbox}
    \caption{Spatiotemporal consistency for 50 sampled data instances, evaluated by both GPT-4 and human judges.}
    \label{tab:gpt4_human_evaluation}
\end{center}
\end{table*}

\section{Further Analyses}
\label{sec:further_analyses}
\subsection{Human Evaluation on Assessing Spatiotemporal Consistency}
We supplement our findings with results from manual evaluations by human judges via Amazon Mechanical Turk on a subset of the dataset used in Table~\ref{tab:main_experiment}.
To specify, we randomly selected 50 instances out of 600, ensuring an even distribution across the four data types (\ie \textit{future}, \textit{past-absence}, \textit{past-presence}, and \textit{past-only} types).
We then applied four methods (\ie zero-shot, RAG-cutoff, narrative-experts, narrative-experts-RAG-cutoff) based on GPT-4 to generate four different responses for each instance.
Subsequently, we tasked annotators with evaluating the spatiotemporal consistency of each response using a binary scale: 0 for inconsistency and 1 for consistency.
Two annotators evaluated each response, and we calculated the average score by dividing the sum of the scores by 2.
Specifically, we opted to assemble a new pool of annotators rather than rely on those who had previously worked on Appendix~\ref{subsec:dataset_quality_details}. This is because we observed that evaluating ‘predicted’ responses posed more challenges than assessing gold responses.
To be similar to Appendix~\ref{subsec:dataset_quality_details}, we collected qualified human annotators via two-step qualification steps, ultimately forming a team of 27 annotators ready to proceed with the main HITs.

The results in Table~\ref{tab:gpt4_human_evaluation} reveal a closely aligned trend between the two evaluation approaches (i.e., GPT-4 evaluation and human evaluation) across methods and data types, with a marginal gap.
In addition, we calculated Gwet’s AC1 scores between two annotators, obtaining a value of \textbf{0.91}.
We also highlight the agreement ratios between annotators who chose the same versus different labels: \textbf{0.94} vs. 0.06 for GPT-4 zero-shot, \textbf{0.94} vs. 0.06 for GPT-4 RAG-cutoff, \textbf{0.96} vs. 0.04 for GPT-4 narrative-experts, \textbf{0.94} vs. 0.06 for GPT-4, and narrative-experts-RAG-cutoff.
This further step emphasizes our commitment to ensuring the reliability and validity of both the automatic evaluation process and our experimental findings.

\subsection{Fine-grained Hallucination Detection}
\begin{table}[t!] \begin{center}
    \begin{adjustbox}{width=\columnwidth}
   \begin{tabular}{lccccc}
       \toprule
           \makecell[l]{Hallucination\\Type} & \makecell{Future} & \makecell{Past-\\absence} & \makecell{Past-\\presence}  & \makecell{Past-\\only} & \makecell{Overall}  \\ %
       \midrule  %
       \makecell[l]{Entity}          & 0/103                      & 0/22                     & 0/8                 & \textbf{24}/66 & 24/199     \\
       \makecell[l]{Relation}          & 0/103                      & 0/22                     & 0/8                 & 2/66 & 2/199     \\
       \makecell[l]{Contradictory}          & 0/103                      & 0/22                     & 1/8                 & 15/66 & 16/199     \\
       \makecell[l]{Invented}          & 0/103                      & 0/22                     & 0/8                 & 14/66 & 14/199     \\
       \makecell[l]{Temporal Error}          & \textbf{103}/103                      & 0/22                     & 0/8                 & 0/66 & \textbf{103}/199     \\
       \makecell[l]{Spatial Error}          & 0/103                      & \textbf{22}/22                     & \textbf{7}/8                 & 0/66 & 29/199     \\
       \makecell[l]{Unclarified}          & 0/103                      & 0/22                     & 0/8                 & 9/66 & 9/199     \\
       \makecell[l]{Incomplete}          & 0/103                      & 0/22                     & 0/8                 & 2/66 & 2/199     \\
       \bottomrule
   \end{tabular}
   \end{adjustbox}
   \caption{Fine-grained hallucination-type distribution from GPT-4 Turbo's zero-shot responses.}
   \label{tab:hallucination_type_distribution}
\end{center}
\end{table}
While the spatiotemporal consistency in Table~\ref{tab:main_experiment} presents only accuracy, GPT-4 judges also provide rationales for their decision.
To thoroughly analyze point-in-time hallucinations in the role-playing LLM, we first use four types of hallucination type from \citet{Mishra:2024:arxiv} and additionally define four types.
\begin{enumerate}
    \item Entity: Incorrect facts about a person, place, or thing.
    \item Relation: Wrong connections between characters or objects.
    \item Contradictory: A statement that goes against known facts.
    \item Invented: Making up facts or details that don't exist.
    \item Temporal error (new): Getting the timing of events wrong, applicable to \textit{future} type instances.
    \item Spatial error (new): Incorrect event participants reasoning, relevant for \textit{past-absence} and \textit{past-presence} type instances. 
    \item Unclarified (new): Not correcting a wrong question, specifically not addressing false information in fake-based questions.
    \item Incomplete (new): Omission of parts of an answer, especially when responding to \textit{free-form} questions.
\end{enumerate}
Then, we collect and manually annotate 199 instances where GPT-4 Turbo's responses are incorrect.

Table~\ref{tab:hallucination_type_distribution} shows a fine-grained hallucination-type distribution of 199 responses.
As anticipated, `Temporal Error' was the most common type, primarily because all the incorrect examples from the \textit{future} type data instances fell under Temporal Error.
On the other hand, most errors in the \textit{past-absence} and \textit{past-presence} were classified as `Spatial Errors'.
The distribution for \textit{past-only} was the most balanced, with a notable number of cases falling into the `Entity' category.
This trend is largely due to instances where role-playing agents failed to detect swapped entities in fake-based questions.

\subsection{Experimental Results per Novel Series}
\begin{table}[t!] \begin{center}
    \begin{adjustbox}{width=\columnwidth}
   \begin{tabular}{lcccccc}
       \toprule
           \makecell[l]{Novel\\Series} & \makecell{Future} & \makecell{Past-\\absence} & \makecell{Past-\\presence}  & \makecell{Past-only\\(Fact)} & \makecell{Past-only\\(Fake)} \\ %
       \midrule  %
       \makecell[l]{Harry Potter}          & 19/50                      & 16/25                     & 24/25                 & 15/25   &  6/25 \\
       \makecell[l]{The Lord of the Rings}          & 27/50                      & 21/25                     & 21/25                 & 23/25  &  17/25 \\
       \makecell[l]{Twilight}          & 23/50                      & 19/25                     & 23/25                 & 18/25   & 5/25 \\
       \makecell[l]{The Hunger Games}          & 24/50                      & 19/25                     & 22/25                 & 18/25  &  16/25 \\
       \bottomrule
   \end{tabular}
   \end{adjustbox}
   \caption{Spatiotemporal consistency of GPT-4 Turbo zero-shot per novel series.}
   \label{tab:experiments_series}
\end{center}
\end{table}
Table~\ref{tab:experiments_series} shows the spatiotemporal consistency of the GPT-4 Turbo zero-shot method across different novel series, highlighting distinct performance trends.
Specifically, the model tends to produce incorrect responses more often for the \textit{future} and \textit{past-absence} types within the Harry Potter series. In contrast, it exhibits lower performance for the \textit{past-presence} type but performs better in other types within the Lord of the Rings series.
Across all series, a common challenge is the model's struggle to accurately respond to instances of both \textit{future} and \textit{past-only (fake)} types.

\subsection{Performance for Fact-Based vs. Fake-Based Questions}
\begin{table}[t!] \begin{center}
    \begin{adjustbox}{width=\columnwidth}
   \begin{tabular}{lcccccc}
       \toprule
        Method & Past-only (Fact) & Past-only (Fake) \\
       \midrule  %
       \multicolumn{3}{l}{\textbf{GPT-4 Turbo} (\texttt{gpt-4-1106-preview})} \\
       \makecell[l]{zero-shot}          & 74.0                      & 44.0 \\
       \makecell[l]{zero-shot-cot}          & 78.0                      & 44.0 \\
       \makecell[l]{few-shot}          & 77.0                      & 57.0 \\
       \makecell[l]{self-refine}          & 82.0                      & 48.0 \\
       \makecell[l]{RAG}          & \textbf{90.0}                      & 54.0 \\
       \makecell[l]{RAG-cutoff}          & 84.0                      & 60.0 \\
       \makecell[l]{\methodName}          & 79.0                      & 56.0 \\
       \makecell[l]{\methodNameWithRag}          & 85.0                      & \textbf{64.0} \\
       \bottomrule
   \end{tabular}
   \end{adjustbox}
   \caption{Spatiotemporal consistency of GPT-4 Turbo zero-shot per fact-based vs. fake-based questions.}
   \label{tab:experiments_fact_vs_fake}
\end{center}
\end{table}
As shown in Table~\ref{tab:experiments_series} and further supported by additional experimental results in Table~\ref{tab:experiments_fact_vs_fake}, LLMs indeed do better at fact-based questions than fake-based ones, as expected. Although fake-based questions might not seem natural within a role-playing scenario, they are crucial from a fact-checking perspective to evaluate the agent’s capability of avoiding point-in-time character hallucinations. Therefore, the inability of the agent to appropriately respond to fake-based questions remains a significant concern.

\subsection{Accuracy of Narrative Experts}
\label{subsec:narrative_experts_accuracy}
While we show the effectiveness of \methodName and \methodNameWithRag, we further analyze the performance of each expert.
To evaluate the temporal expert, we use all 600 data instances used in Table~\ref{tab:main_experiment}, since all data instances are classified as either future or past.
On the other hand, we only use 200 \textit{fact-based structured} data instances from \textit{past-absence} and \textit{past-presence} types to evaluate the spatial expert.

\begin{table}[t!] \begin{center}
    \begin{adjustbox}{width=\linewidth}
   \begin{tabular}{lcccc}
       \toprule
           \multirow{2}{*}{\makecell[l]{Method}}  & \multicolumn{2}{c}{Temporal Expert} & \multicolumn{2}{c}{Spatial Expert} \\
           \cmidrule(r{0.3em}){2-3}
           \cmidrule(r{0.3em}){4-5}
           & \makecell{Future} & \makecell{Past} & \makecell{Absence}  & \makecell{Presence}  \\ %
       \midrule  %
       Random & 50.0 & 50.0 & 50.0 & 50.0 \\
       \midrule
       \multicolumn{5}{l}{\textbf{Mistral Instruct 7B} (\texttt{mistral-7b-instruct-v0.2})}     \\
       \makecell[l]{\methodName}          & 30.5                      & 81.0                     & 70.0                 & 87.0      \\
       \makecell[l]{\methodNameWithRag}          & 63.5                      & 89.3                     & 76.0                 & 88.0      \\
       \midrule
       \multicolumn{5}{l}{\textbf{GPT-3.5 Turbo} (\texttt{gpt-3.5-turbo-1106})}     \\
       \makecell[l]{\methodName}          & 75.5                      & 79.8                     & 68.0                 & 90.0      \\
       \makecell[l]{\methodNameWithRag}          & \textbf{89.5}                      & 76.8                     & 68.0                 & 91.0      \\
       \midrule
       \multicolumn{5}{l}{\textbf{GPT-4 Turbo} (\texttt{gpt-4-1106-preview})}     \\
       \makecell[l]{\methodName}          & 83.5                      & \textbf{94.5}                     & 76.0                 & \textbf{98.0}      \\
       \makecell[l]{\methodNameWithRag}          & 85.0                      & 94.3                     & \textbf{84.0}                 & 96.0      \\
       \bottomrule
   \end{tabular}
   \end{adjustbox}
   \caption{Accuracy of temporal and spatial experts.}
   \label{tab:narrative_experts_accuracy}
\end{center}
\end{table}

Table~\ref{tab:narrative_experts_accuracy} demonstrates the accuracy of narrative experts using three backbone LLMs.

\textbf{\methodName \textit{vs.} \methodNameWithRag.}
Despite lacking access to external knowledge, narrative experts significantly outperform random selection in classifying past/future and presence/absence scenarios.
This indicates that LLMs rely on their parametric memories for spatiotemporal reasoning. 
Moreover, \methodNameWithRag outperforms \methodName on average, highlighting the benefits of incorporating retrieval modules. 

\textbf{Comparison of backbone models.}
As expected, narrative experts using GPT-4 Turbo achieve the highest accuracy on average.
While the accuracy of temporal experts with both GPT-4 Turbo and GPT-3.5 Turbo in predicting future events exceeds 75\%, the temporal expert with Mistral faces challenges in future predictions.
Conversely, Mistral surpasses GPT-3.5 Turbo in predicting past events and absences.

\textbf{Instruction following capability.}
The temporal expert of GPT-3.5 Turbo \methodNameWithRag shows an impressive 89.5\% accuracy in detecting future events. 
However, its performance on the \textit{future} type in Table~\ref{tab:main_experiment} reveals that the GPT-3.5 Turbo model, when acting as a role-playing agent, often fails to follow provided instructions.
Furthermore, we demonstrate that the underperformance of \methodName and \methodNameWithRag in the \textit{past-presence} type, as shown in Table~\ref{tab:main_experiment}, is often due to the temporal and spatial experts' occasional failures to predict past events and presences accurately.
Consequently, this leads to the role-playing LLMs following `incorrect' instructions.

\subsection{Evaluating Role-Playing LLM Agents on Time Points More Distant in the Future}
While we presented Figure~\ref{fig:point_in_time_character_hallucination} as a motivational example of \datasetName, we observed that the challenge of detecting character hallucination, particularly with \textit{future} type questions, diminishes as the temporal gap (\ie the gap between the character’s time point and the question's time point) widens.
Conversely, it remains challenging for LLMs to identify hallucinations when the temporal difference narrows, as illustrated in Table~\ref{tab:main_experiment}.
To empirically validate this observation, we designed an experiment as follows: We selected 15 questions from the latter part of ``Harry Potter and the Deathly Hallows'' (book 7), allocating them between Harry Potter and Hermione Granger - 8 for Harry and 7 for Hermione.
We then positioned the characters at seven distinct time points (\ie the start of each school year from year 1 to year 7), resulting in a total of $15 * 7 = 105$ unique \textit{future} type instances.
Note that this scalable approach enabled us to generate examples that assess models at time points distant in the future.

\begin{table}[t!] \begin{center}
    \begin{adjustbox}{width=\columnwidth}
   \begin{tabular}{lccccccc}
       \toprule
           \makecell[l]{Character} & 6 years & 5 years & 4 years & 3 years & 2 years & 1 years & 0 year  \\
        \midrule
            \multicolumn{7}{l}{\textbf{GPT-3.5 Turbo} (\texttt{gpt-3.5-turbo-1106}) zero-shot} \\
            \makecell[l]{Harry Potter}          & 25.0 & 25.0 & 0.0 & 12.5 & 12.5 & 12.5 & 0.0 \\
            \makecell[l]{Hermione Granger}          & 71.4 & 57.1 & 42.9 & 42.9 & 42.9 & 42.9 & 14.3 \\
            \makecell[l]{\textbf{Total}}          & 46.7 & 40.0 & 20.0 & 26.7 & 26.7 & 26.7 & 6.7 \\
        \midrule
            \multicolumn{7}{l}{\textbf{GPT-4 Turbo} (\texttt{gpt-4-1106-preview}) zero-shot} \\
            \makecell[l]{Harry Potter}          & 100.0 & 100.0 & 100.0 & 100.0 & 100.0 & 100.0 & 25.0 \\
            \makecell[l]{Hermione Granger}          & 100.0 & 100.0 & 100.0 & 100.0 & 85.7 & 85.7 & 71.4 \\
            \makecell[l]{\textbf{Total}}          & 100.0 & 100.0 & 100.0 & 93.3 & 93.3 & 93.3 & 46.7 \\
       \bottomrule  %
       \end{tabular}
   \end{adjustbox}
   \caption{Spatiotemporal consistency across \textit{future} type instances, divided by the temporal gap between the role-playing character's time point and the question's time point.}
   \label{tab:distant_future_experiment}
\end{center}
\end{table}

The Table~\ref{tab:distant_future_experiment} presents the spatiotemporal consistency across 105 instances, illustrating the effect of the temporal gap between the character's time point and the question’s time point on model performance.
According to the results, LLMs (with zero-shot prompt) exhibit an increase in spatiotemporal consistency as the temporal distance between the character’s time point and the question’s time point widens.
Specifically, `GPT-3.5 zero-shot' showed a maximum of 46.7\% spatiotemporal consistency, indicating persistent hallucinations, while `GPT-4 zero-shot' achieved 100\%.
This suggests that role-playing LLMs, especially GPT-4, with a higher memorization capability, are able to recognize this extent of temporal difference as the temporal gap between the character and the question expands in \textit{future} type instances.
However, it remains challenging for LLMs to maintain spatiotemporal consistency when the temporal distance between the character’s time point and the question’s time point decreases (e.g., GPT-3.5: 6.7\%, GPT-4: 46.7\%), as also depicted in Table~\ref{tab:main_experiment}.

\end{document}